\documentclass[sigconf]{acmart}

\usepackage{amsmath,amssymb,amsfonts}
\usepackage{algorithmic}
\usepackage{graphicx}
\usepackage{graphbox}
\usepackage{tabularx}
\usepackage{booktabs}
\usepackage{textcomp}
\usepackage{xcolor}
\usepackage{courier}
\usepackage{subcaption}
\usepackage{multirow}
\usepackage{enumitem}
\usepackage{url}
\usepackage{soul}

\def\BibTeX{{\rm B\kern-.05em{\sc i\kern-.025em b}\kern-.08em
    T\kern-.1667em\lower.7ex\hbox{E}\kern-.125emX}}

\begin{document}

\title{MalLight: Influence-Aware Coordinated Traffic Signal Control for Traffic Signal Malfunctions}

\author{Qinchen Yang}
\affiliation{%
  \institution{Rutgers University}
  \city{Piscataway}
  \state{NJ}
  \country{USA}}
\email{qy129@rutgers.edu}

\author{Zejun Xie}
\affiliation{%
    \institution{Rutgers University}
    \city{Piscataway}
    \state{NJ}
    \country{USA}}
\email{zx180@scarletmail.rutgers.edu}

\author{Hua Wei}
\affiliation{%
    \institution{Arizona State University}
    \city{Tempe}
    \state{AZ}
    \country{USA}}
\email{hua.wei@asu.edu}

\author{Desheng Zhang}
\affiliation{%
  \institution{Rutgers University}
  \city{Piscataway}
  \state{NJ}
  \country{USA}}
\email{desheng@cs.rutgers.edu}

\author{Yu Yang}
\affiliation{%
  \institution{Lehigh University}
  \state{Bethlehem}
  \state{PA}
  \country{USA}}
\email{yuyang@lehigh.edu}

\begin{abstract}
Urban traffic is subject to disruptions that cause extended waiting time and safety issues at signalized intersections. While numerous studies have addressed the issue of intelligent traffic systems in the context of various disturbances, traffic signal malfunction, a common real-world occurrence with significant repercussions, has received comparatively limited attention. The primary objective of this research is to mitigate the adverse effects of traffic signal malfunction, such as traffic congestion and collision, by optimizing the control of neighboring functioning signals. To achieve this goal, this paper presents a novel traffic signal control framework (MalLight), which leverages an Influence-aware State Aggregation Module (ISAM) and an Influence-aware Reward Aggregation Module (IRAM) to achieve coordinated control of surrounding traffic signals. To the best of our knowledge, this study pioneers the application of a Reinforcement Learning(RL)-based approach to address the challenges posed by traffic signal malfunction. Empirical investigations conducted on real-world datasets substantiate the superior performance of our proposed methodology over conventional and deep learning-based alternatives in the presence of signal malfunction, with reduction of throughput alleviated by as much as 48.6$\%$.
\end{abstract}



\maketitle

\vspace{-10pt}
\section{Introduction}
Traffic signal malfunction refers to a situation when a traffic signal, also known as a traffic light, fails to function properly.
One of the typical situations is that traffic lights are blackout caused by various reasons, such as flood~\cite{lhomme2013analyzing,suarez2005impacts}, hurricane~\cite{pregnolato2016assessing}, equipment aging~\cite{cools2013self}, or human-made events, such as cyber-attacks~\cite{lee2019cyber,ghena2014green}, traffic accidents and mismanagement~\cite{chowdhury2013traffic}.

\begin{figure}[ht]
\setlength{\belowcaptionskip}{-0.8cm}
\centering
\begin{minipage}[t]{0.48\linewidth}
        \centering
        \includegraphics[width=4.2cm]{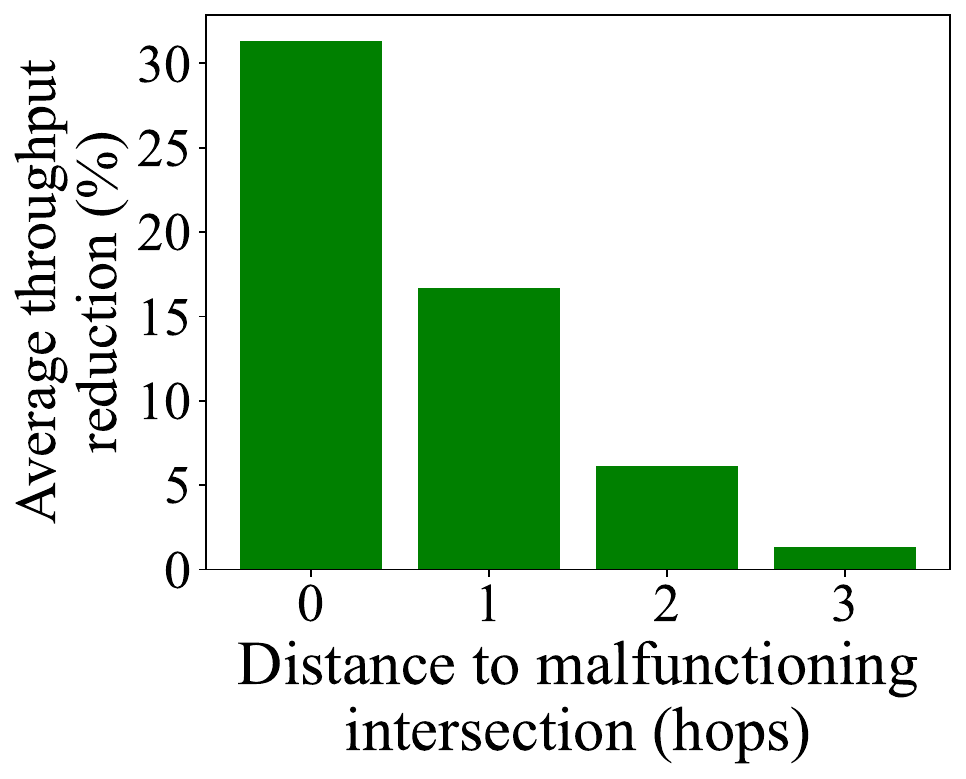}
        \caption{Throughput reduction at intersections around malfunctioning traffic signal}
        \label{fig:figure-RR}
    \end{minipage}
    \hspace{0.1cm}
    \begin{minipage}[t]{0.48\linewidth}
        \centering
        \includegraphics[width=4.2cm]{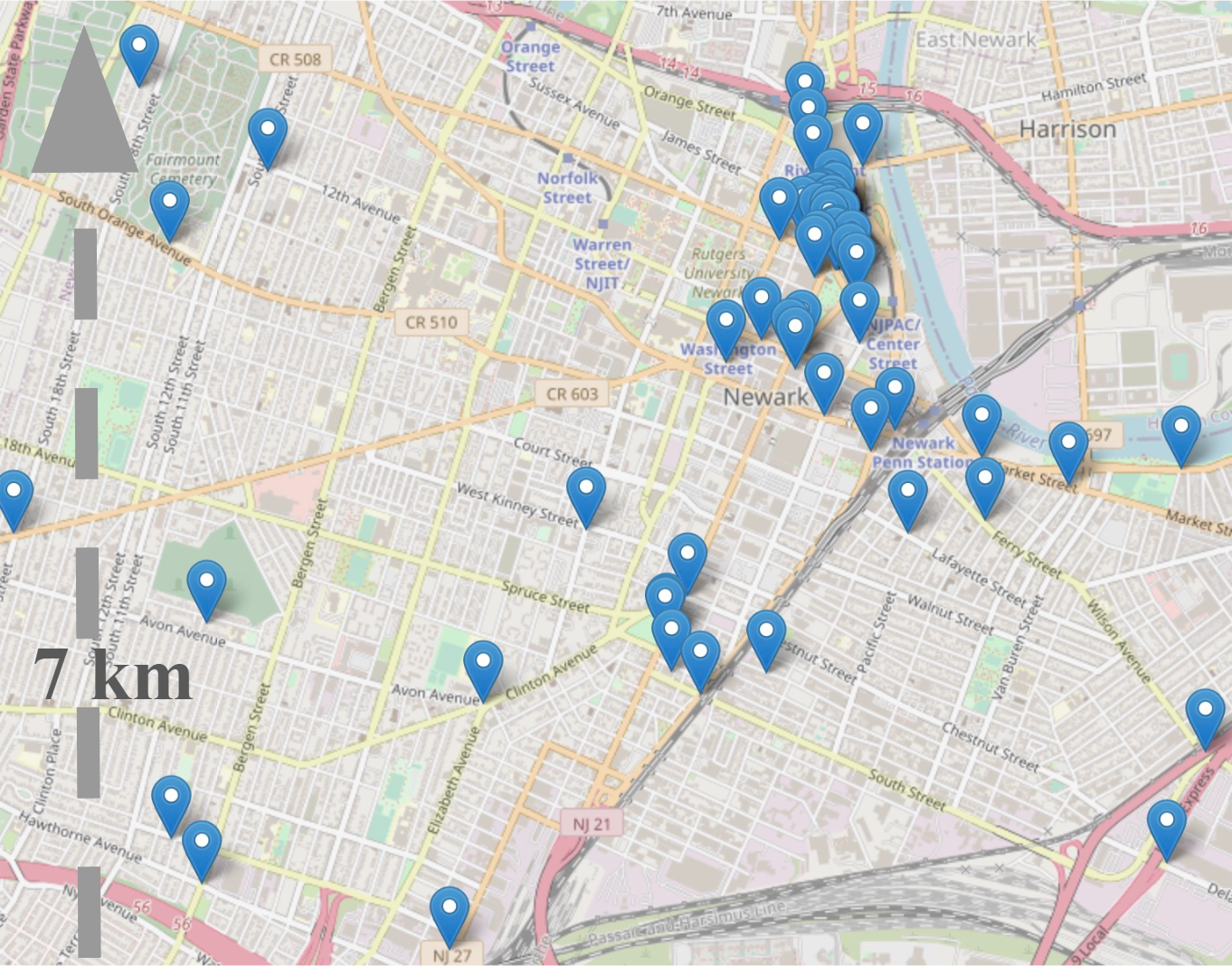}
        \caption{Locations of complaints about traffic signal malfunction}
        \label{fig:figure-mal}
    \end{minipage}
\end{figure}

When the traffic signal malfunction happens, it causes confusion and chaos on the road, leading to increased risk of accidents, traffic delay,s and traffic congestion.
For example, we conducted a case study based on the traffic dataset from New York City (details in the Evaluation section), where traffic signals were set to be controlled by the FixedTime strategy~\cite{koonce2008traffic}, the most commonly used traffic signal control strategy.
When the traffic signal at one intersection experiences a blackout, the average throughput of that intersection is reduced by nearly 30\%.
Moreover, the impacts also propagated to nearby intersections, such as intersections 1, 2, and 3-block away from the malfunctioning intersection, as shown in Figure~\ref{fig:figure-RR}.
Similar observations are identified in other studies with increased travel time~\cite{ezell2013cyber} and fatal accidents~\cite{NCSA2022}.
More importantly, these malfunctions occur more frequently than one might expect.
For example, as shown in Figure~\ref{fig:figure-mal}, citizens from the City of Newark in New Jersey reported 94 complaints about traffic signal malfunctions at 67 different intersections between August 2020 and May 2022 on the Newark Connect platform~\cite{newarkconnect} (i.e., a 311-like platform for citizens to report city issues), leading to three issues per month on average.
Consequently, it is necessary to find a way that can mitigate the negative impacts (e.g., congestion) when traffic signal malfunctions occur.

The problem of traffic signal malfunctions has been studied in existing work~\cite{chen2009assessment,purba2018developing,hunter2011mode}.
In general, we can categorize these studies into two stages: detection and response.
The malfunction detection mainly relies on road user report~\cite{chen2009assessment} and some recently developed intelligent signal systems to automatically detect signal malfunction~\cite{soh2013smart,purba2018developing}.
After detection, the ultimate response is to repair or replace the malfunctioning traffic signals.
Unfortunately, due to limited resources (e.g., human resources and budgets), this response process can be time-consuming, with repairs sometimes taking months or even years to complete~\cite{chicago}.
Therefore, immediate temporary responses become necessary.
Current temporary responses heavily rely on manual efforts, such as manual traffic guidance or Flash Mode~\cite{benioff1980study}.
However, abundant ground surveys have shown that Flash Mode is less effective~\cite{hunter2011mode}, and manual traffic guidance also suffers from long time delays~\cite{al2003dynamic,taale2004evaluation}.
Our work aims to provide a faster and more automated response after the malfunctions are detected and before human efforts get involved.

Our solution is inspired by the fundamental theory of traffic management from the perspective of demand (i.e., the number of vehicles that request to pass through an intersection in a unit of time) and capacity (i.e., the number of vehicles that an intersection can handle in the same unit of time)~\cite{aftabuzzaman2007measuring,rosenbloom1978peak}.
Traffic congestion occurs when demand is higher than capacity.
When a traffic signal malfunction occurs, i.e. the intersection changes from signalized to a blackout, capacity decreases~\cite{paul2017comparison}, while demand remains relatively constant in the short term.
This imbalanced relationship between demand and capacity motivates us to explore immediate responses to traffic signal malfunction:
by carefully managing the networked traffic signals, especially those surrounding malfunctioning intersections, we can potentially control the traffic demand that enters the malfunctioning intersections in a desired manner, which leads to a more balanced demand and capacity.

However, coordinating intersections to control traffic in a desired manner is challenging, because (1) the impact from each intersection to malfunctioning ones is hard to measure. For example, one intersection has several routes to malfunctioning intersections. (2) Influence changes over time, such as during peak hours and non-peak hours.

Recently, reinforcement learning (RL) has introduced a new possibility to coordinate traffic signal controls (TSC) ~\cite{chen2020toward,du2023safelight,wei2019colight,presslight19,zheng2019learning}, where each agent controlling a single intersection can automatically learn to generate its actions to maximize its own benefits (e.g., the throughput of corresponding intersection).
Since the reward function for each agent in these RL methods only depends on its own traffic condition, under a malfunction scenario, the agents of the well-functioning intersections are not aware of the specific needs of the malfunctioning intersections, even when its assistance benefits the whole system. Through experiments (Table~\ref{table-result}, we observe that previous RL methods experience significant throughput reduction. For example, the IDQN method experiences a 42.8$\%$ reduction in throughput on the Hangzhou dataset, which is roughly 3.3 times higher than the 13$\%$ reduction observed with our model.
Further, how the actions of each agent impact others is not well studied.
To the best of our knowledge, none of the existing RL-based TSC methods can be applied in malfunction setting.

In this work, we design MalLight, an RL-based traffic signal control method to mitigate the impacts of traffic signal malfunctions, by modeling the spatially and temporally varied influence between well-functioning and malfunctioning interactions. Specifically, MalLight models the dynamic influences through the special design of the state module and the reward module in RL. 
To model the spatially and temporally varied influence, we adapt a graph diffusion convolution network to integrate the network-level state with the local state of an individual agent so that each RL agent is aware of its influence on other intersections while controlling its local traffic conditions.
To guide each well-functioning intersection to be aware of the reward associated with malfunctioning intersections, we introduce the graph diffusion process. This process enables each agent to dynamically integrate rewards from multiple malfunctioning intersections while learning their policy.
In summary, the main contributions of this work are as follows:
\\\noindent$\bullet$~This is the first work that utilizes adaptive traffic signal control to build a resilient and responsive traffic system in the presence of traffic signal malfunction.
\\\noindent$\bullet$~We design a novel RL model, MalLight, based on modeling influences among intersections under signal malfunction. 
We incorporate the graph diffusion process into the state and reward design to model the influences and guide the training of MalLight, making it resistant to signal malfunction.
\\\noindent$\bullet$~Experiments on real-world and synthetic datasets show that MalLight outperforms both traditional and state-of-the-art (SoTA) RL-based models, achieving a remarkable alleviation in throughput reduction by up to 48.6\%.
\vspace{-10pt}
\section{Preliminaries}
In this section, we introduce the preliminaries of the traffic signal control problem~\cite{wei2019presslight, wei2019colight} and extend it to our scenario considering traffic signal malfunction.
In our setting, each intersection is controlled by one agent, which has its own local observations of its intersection as the state.
An agent controls the traffic signal in its own intersection by deciding the signal phase at the next time interval to optimize the local traffic, such as maximizing the throughput.
We use centralized learning with a decentralized execution manner to reduce the computational complexity of traditional centralized methods~\cite{wei2019survey}.
Specifically, we formulate our problem as a Markov Decision Process(MDP) with the following components $<S,O,M,A,r,\pi,\gamma>$.

\noindent$\bullet$~\textbf{\textit{Set of malfunctioning signals $M$}}: 
$M$ is a set of intersections with malfunctioning traffic signals.

\begin{figure*}[t]
    \centering
    \begin{minipage}{0.25\textwidth}
        \centering
        \includegraphics[height=3.3cm]{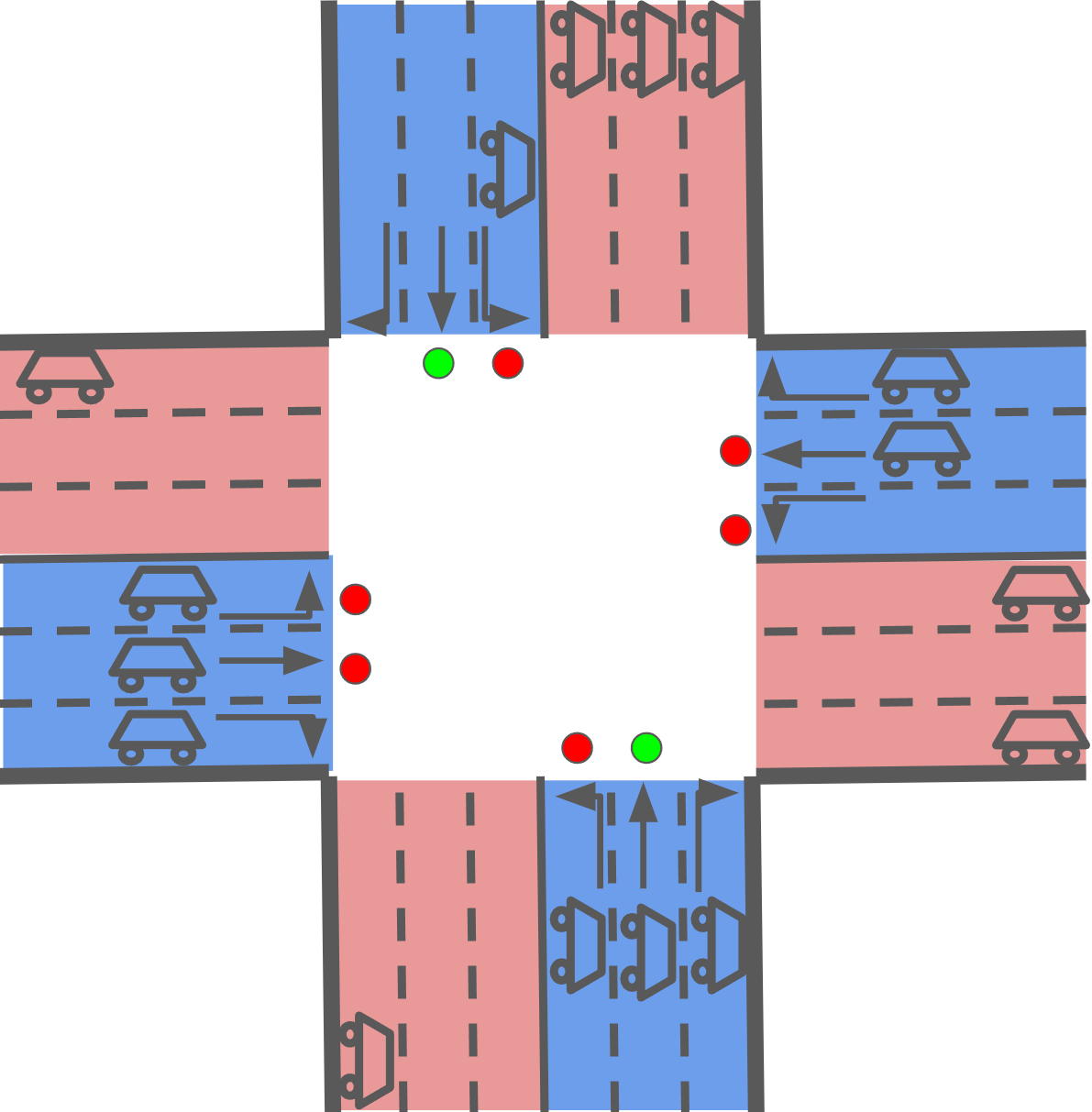}
        \caption{Intersection, incoming lanes(blue), outgoing lanes(red). Phase 3 is set.}
        \label{figs:intersection}
        \includegraphics[height=1.9cm]{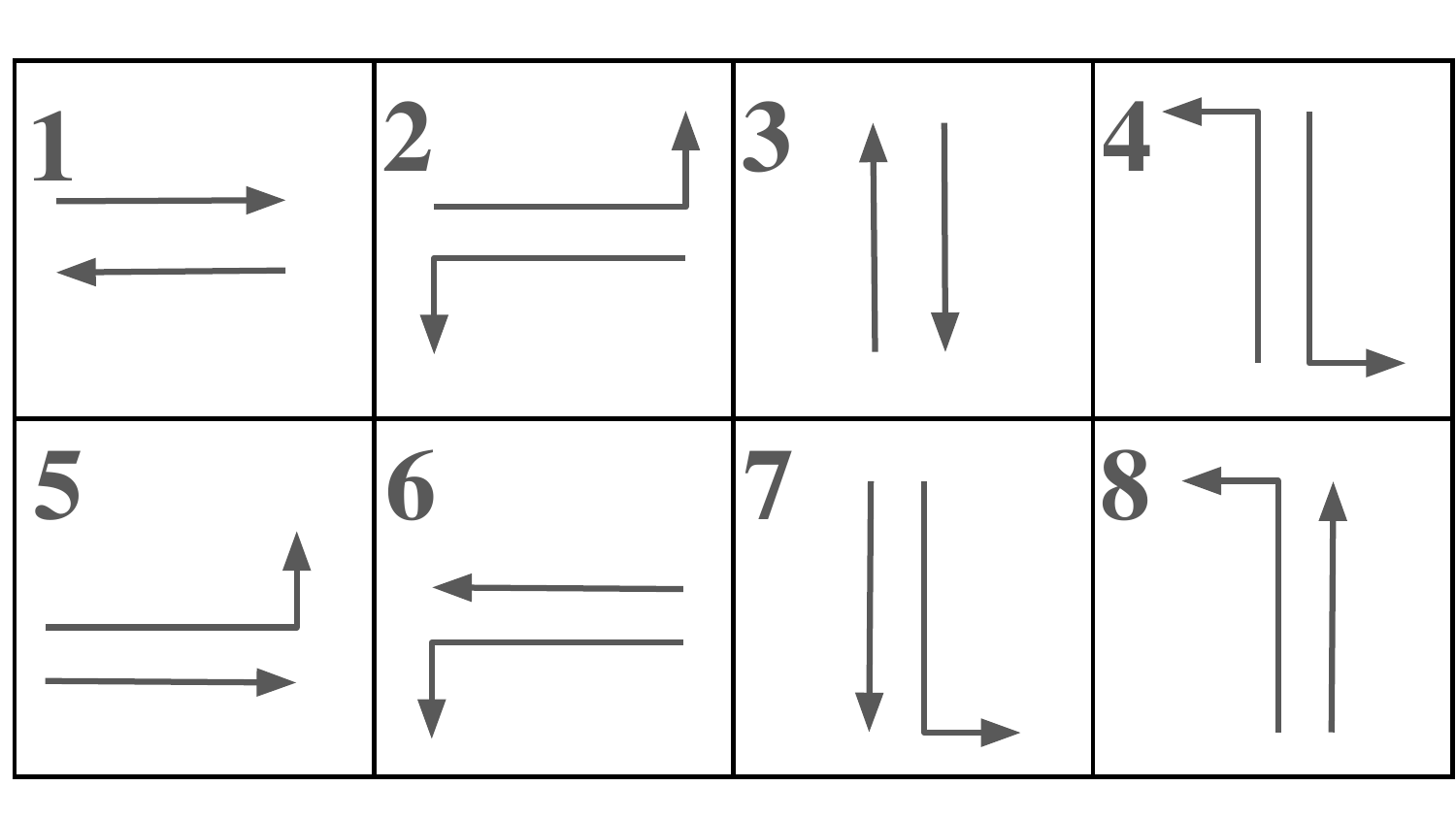}
        \caption{Eight phases (actions).}
        \label{figs:actions}
    \end{minipage}
    \hspace{0.1cm}
    \begin{minipage}{0.7\textwidth}
    \setlength{\belowcaptionskip}{-15pt}
        \centering
        \includegraphics[width=0.99\linewidth]{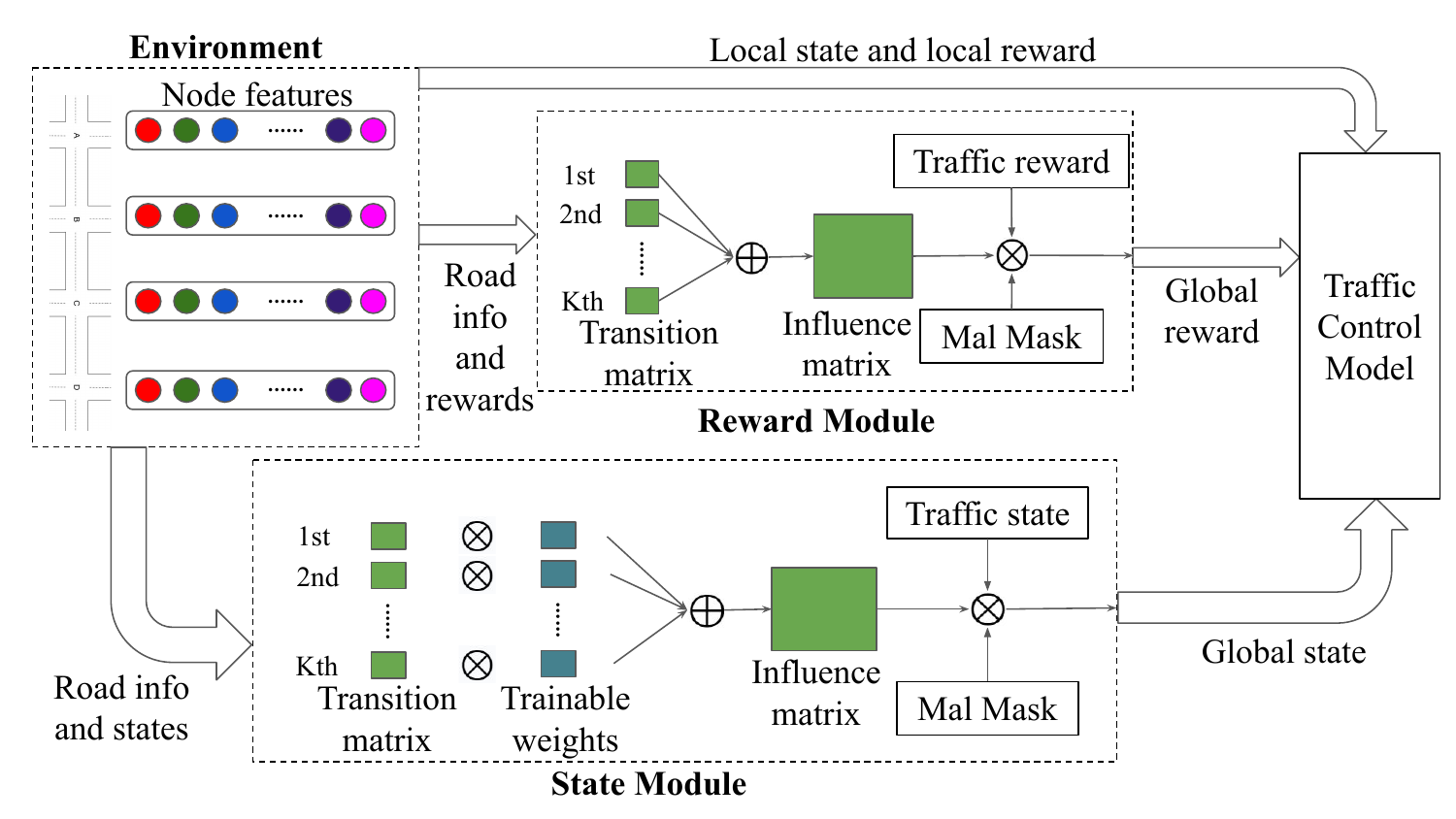}
        \vspace{-15pt}
        \caption{MalLight framework and modules.}
        \label{figure-model}
    \end{minipage}
\end{figure*}

\noindent$\bullet$~\textbf{\textit{State space $S$ and observation space $O$}}: 
Assuming there are $N$ intersections in our road network and $N$ corresponding agents that control the traffic signals located at the intersections. 
All possible traffic states from $N$ intersections form the state space $S$ and each agent has access to partial state s$\in$$S$, which forms the agent's own observation o$\in$$O$. 
The state $s$ of each intersection includes current phase $p$ (one of the 8 total phases illustrated in Figure~\ref{figs:actions}) and the number of vehicles in each lane (i.e., a 12-element vector with each element representing a number of vehicles in lanes illustrated in Figure~\ref{figs:intersection}).
In MalLight, observations from all intersections in the road network are fed into a state module, which outputs a global state. 
The global state is combined with local state $s$ to form observation $o$ for each agent.

\noindent$\bullet$~\textbf{\textit{Action space $A$}}: 
In the traffic signal control task, the $i$th agent generates an action $a_i^t$ as the signal phase at time point $t$ from its action space $A_i$ for the following $\Delta$$t$ period. Eight signal phases corresponding to eight actions are illustrated in Figure~\ref{figs:actions}. The eight-phase setting is commonly used in real-world traffic signal control task~\cite{wei2019survey}.
We set $\Delta$$t$ as 10 seconds based on the practice of the previous work~\cite{chen2020toward}.
Given the action $a_i^t$, the traffic signal chooses the corresponding phase $p$. 
For malfunctioning signal $m \in M$, $a_m^t$ is set to be off denoted as $p_{mal}$.

\noindent$\bullet$~\textbf{\textit{Reward $r$}}: 
Each RL agent obtains an immediate reward from the environment at time $t$ by a rewarding function $S \times A_1 \times ... \times A_N \to R$, where $N$ is the number of intersections.
We introduce the concept of \textit{pressure} to measure the local reward for each intersection, which is the difference between the upstream and downstream queue length, indicating the inequivalence of the vehicle distribution~\cite{chen2020toward}.
More specifically, we denote the local reward of the $i$th intersection at time $t$ as $r_i^t$, which is calculated as
\begin{equation}
\setlength{\abovedisplayskip}{1pt}
\setlength{\belowdisplayskip}{1pt}
r_i^t = -P_i = -\left|\sum_{l_{in}\in{i}}u^t_{i,l_{in}} - \sum_{l_{out}\in{i}}u^t_{i,l_{out}}\right|
\end{equation}
where $P_i$ represents the pressure of intersection $i$, $u^t_{i,l}$ is the queue length of lane $l$ at intersection $i$ at time $t$, $l_{in}$ and $l_{out}$ represent incoming and outgoing lane, respectively.
For example, in Figure~\ref{figs:intersection}, the pressure of intersection is calculated as $|$\#queueing cars in incoming lanes - \#queueing cars in outgoing lanes$|$, which is $|3+3+1+2-1-1-3-2|=2$.
In MalLight, rewards from all intersections in the road network are fed into a reward module, which generates a global reward. Details of reward generation are described in Section~\ref{section:rewardAggregation}.
The global reward is then combined with the local reward $r_i^t$, to form the final reward $r_i^{,t}$ for each agent.

\noindent$\bullet$~\textbf{\textit{Policy set $\pi$ and discount factor $\gamma$}}: 
At time $t$, each agent chooses an action following a policy $\pi$:$S\to{A}$. 
This choice can also be expressed as action-value function $Q^\pi$, which indicates the expected quality of an action $a_i^t$ if the $i$th agent takes action $a_i^t$ based on policy $\pi$ in state $s_t^i$. 
The formulation of the action-value function is
\begin{equation}
\setlength{\abovedisplayskip}{1pt}
\setlength{\belowdisplayskip}{1pt}
    Q^\pi(s,a)=\mathbb{E}[\sum_{i=t}^{\infty}{\gamma r_{t+i}|s_t^i=s,a_t^i=a}]
\end{equation}

\noindent\textbf{\textit{Problem formulation and objective}}:
In our problem, each intersection is controlled by one agent, and these agents share parameters from the same deep RL model.
At each time step $t$, agent $i$ (managing well-functioning traffic signals) obtains an observation $o_i^t$ from the environment.
Given traffic conditions and current signal phases at both local and other intersections, the objective of each agent is to determine the optimal action $a_i^t$ that maximizes the reward $r_i^{,t}$, formulated as:
\begin{equation}
\setlength{\abovedisplayskip}{1pt}
\setlength{\belowdisplayskip}{1pt}
arg\max_{a_i^t\in A} r_i^{,t} = -P_i - \sum_{j\in M}(w_j P_j)
\end{equation}
Here, $P_i$ represents the pressure at $i$th intersection, $w_j$ is the weight of $j$th intersection with malfunctioning signals, for reward aggregation, and $M$ is the set of intersections of malfunctioning signals.
\section{Method}
We introduce MalLight, a RL-based traffic signal control model built upon a general RL framework, incorporating new modules: an Influence-aware State Aggregation Module (ISAM) and an Influence-aware Reward Aggregation Module (IRAM). 
These modules enhance the model's resistance to signal malfunctions.
In the following sections, we first introduce the overview of our model and then describe the detailed design of ISAM and IRAM.
\subsection{Overview}
To alleviate congestion in malfunctioning intersections through coordinated management with neighboring intersections (intersections can be multi-hop away), each agent should possess two key abilities. 
Firstly, each agent should have the ability to ``care'' about malfunctioning intersections. 
In previous works~\cite{wei2019colight,wei2019presslight,chen2020toward,zhang2022expression}, the reward for each RL agent was solely dependent on its own traffic conditions, which meant that each agent would only benefit from its own actions.
In our setting, intercepting some vehicles at a well-functioning intersection to prevent excessive traffic from entering malfunctioning intersections would lead to a reduction in the rewards earned by agents at well-functioning intersections.
Therefore, to alleviate the traffic demand on malfunctioning intersections, each surrounding agent should be aware of the reward from these malfunctioning intersections.
Secondly, each agent should have the ability to quantify how much influence it can exert on the malfunctioning intersection to determine the optimal actions.
Different levels of influence require different policies.

The overall framework of MalLight is depicted in Figure~\ref{figure-model}.
In line with the two aforementioned abilities, we introduce two modules, the Influence-aware State Aggregation module (ISAM), and the Influence-aware Reward Aggregation module (IRAM).

In ISAM, we designed masked graph diffusion convolution and adapt it into our multi-agent scenario, where each agent needs to take nearby malfunctioning agents into consideration while being aware of its influence on these agents. In IRAM, we design a novel reward reshaping scheme based on diffusion process, which boost the performance of model when used in cooperation with masked diffusion convolution. Up to our knowledge, we are the first to adapt diffusion convolution into multi-agent RL scheme and the first to reshape reward by diffusion process.

To simplify the task, in MalLight, sensing devices for detecting the vehicle queue length of each lane, such as cameras or inductive loop detectors, are assumed to work well in all intersections including the malfunctioning ones, as traffic signals and sensing devices are generally managed by different systems~\cite{chen2009assessment,kuppusamy2018design}.
In addition, we assume that the system is aware of which intersections have malfunctioning traffic signals (e.g., through human reports or anomaly detection techniques~\cite{purba2018developing,soh2013smart}), and the signal control agents can share information.

\vspace{-5pt}
\subsection{Influence-aware State Aggregation Module}
The Influence-aware State Aggregation Module (ISAM) is designed to ensure that each agent's observation includes global states from malfunctioning intersections while taking into account its influence on these malfunctioning intersections. 
Higher influence corresponds to greater weight assigned to other intersections' states, and vice versa.
ISAM takes road network static information and intersection-specific local states as input. 
For each agent, it generates an aggregated state that encompasses observations from malfunctioning intersections and their corresponding local states. 
This state information is subsequently fed into the RL model and serves as the agent's final environmental observation.

Inspired by the graph diffusion convolutional recurrent neural network (DCRNN)~\cite{li2017diffusion}, which has achieved promising results in automatically modeling complex spatial dependencies on road networks, we design masked diffusion convolution and adapt it to multi-agent scheme, to capture influence among intersections, particularly the influence from well-functioning intersections to malfunctioning intersections.
In the diffusion convolution operation, an influence vector is extracted for each agent. These vectors are used as weights to perform a weighted sum of states from other intersections. Additionally, before the weighted sum is calculated, a Malfunction Mask is applied to filter out data from well-functioning intersections. 
This helps agents at well-functioning intersections focus on only themselves and the malfunctioning intersections.
Details of ISAM are described below.

\noindent$\bullet$~\textbf{\textit{Diffusion Process.}} Diffusion convolution~\cite{li2017diffusion} is a variation of the diffusion process~\cite{teng2016scalable} that incorporates trainable parameters into transition matrix of each layer, offering enhanced representativeness and flexibility in modeling influence.
According to~\cite{teng2016scalable}, the stationary distribution of the diffusion process can be expressed as a weighted combination of random walks on the graph:
\begin{equation}
\setlength{\abovedisplayskip}{1pt}
\setlength{\belowdisplayskip}{1pt}
    \mathcal{P}=\sum_{k=1}^{K} \alpha(1-\alpha)^k(D_O^{-1}W)^k,
\end{equation}
, where $k$ represents the diffusion step, $W$ is the edge weight matrix, $D_O$ is the out-degree diagonal matrix of graph, $\alpha\in[0,1]$ denotes the restart probability, and $D_O^{-1}W$ serves as the state transition matrix. 
This Markov process converges to the distribution $\mathcal{P}\in \mathbb{R}^{N\times N}$, where each $i$-th row $\mathcal{P}_i,:$ represents the likelihood of diffusion from node $v_i\in V$, where $V$ represents the set of nodes in the graph, and $N$ denotes the total number of nodes.

The edge weight matrix $W$ is constructed based on the road distance between each pair of intersections.
For normalization, we use a thresholded Gaussian kernel~\cite{shuman2013emerging}, defined as follows:
\begin{equation}
\setlength{\abovedisplayskip}{1pt}
\setlength{\belowdisplayskip}{1pt}
    W_{i,j}=exp(-\frac{dist(v_i,v_j)^2}{\sigma^2})
\end{equation}
In this formula, $W_{i,j}$ represents the edge weight between node $v_i$ and node $v_j$, $\sigma$ is the deviation of distances, and $dist(v_i,v_j)$ is 0 if nodes $v_i$ and $v_j$ are not connected by a road. Otherwise, it represents the distance between $v_i$ and $v_j$.

\noindent$\bullet$~\textbf{\textit{Masked Diffusion Convolution.}} Building upon the diffusion process described above, diffusion convolution introduces a trainable filter for each diffusion step $k$. Additionally, to ensure that agents do not concentrate their efforts on other functioning intersections, which do not require assistance, we apply the Malfunction Mask.
The Malfunction Mask vector, denoted as $Mask\in \{0,1\}^N$, where $N$ represents the number of intersections in the road network. 
The element in the Malfunction Mask is set to 1 if the corresponding intersection is experiencing a malfunction, and 0 otherwise.
The formulation of masked diffusion convolution is as follows:
\begin{equation}
\setlength{\abovedisplayskip}{1pt}
\setlength{\belowdisplayskip}{1pt}
    S'_{:,p}=\left[\sum_{k=1}^{K}(\theta_k(D_O^{-1}W)^k) \odot Mask\right] S_{:,p}, p\in\{1,2,3,...,P\}
\end{equation}
Here, $\odot$ indicates Hadamard product, $S\in \mathbb{R}^{N\times P}$ represents the matrix of features in the graph (each agent's local state), $P$ is the length of input features, $p\in [1,P]$ is an index of an element in input feature, $\theta_k$ represents the trainable filter for diffusion step $k$, and $K$ is the number of diffusion steps.
The masked diffusion convolution process involves random walks on the road network for each element of the input features. 
$S'_{i,:}$ indicates the information i-th agent needs considering its influence to other nodes(intersections).
Masked diffusion convolution provides a more explainable method for making the model aware of the influence.

\noindent$\bullet$~\textbf{\textit{State Aggregation.}}
This step combines the local state of the target intersection with the global state $S'$ originating from malfunctioning intersections. For the $i$-th agent operating at the $i$-th intersection, its corresponding global state is $S'_{i,:}\in \mathbb{R}^P$ and its local state is $S_{i,:}\in \mathbb{R}^P$. Finally, for the $i$-th agent, we obtain the state as:
\begin{equation}
\setlength{\abovedisplayskip}{1pt}
\setlength{\belowdisplayskip}{1pt}
    S''_{i,:} = S'_{i,:} + S_{i,:}, i\in \{1,2,3,...,N\}
\end{equation}
Here $S''_{i,:}\in \mathbb{R}^{P}$ contains all the information i-th agent should consider for making a decision and it serves as the input to the traffic control RL model.

\subsection{Influence-aware Reward Aggregation Module}\label{section:rewardAggregation}
The Influence-aware Reward Aggregation Module (IRAM) is designed to ensure that each agent ''cares'' about the benefits of malfunctioning intersections and is ''willing'' to assist these intersections.
IRAM takes road network static information and intersection rewards as input, and for each agent, generate aggregated reward containing rewards of malfunctioning intersections and corresponding local rewards. The reward information then serve as direction to optimize the RL model.

Reward Shaping~\cite{ng1999policy} is a technique, where additional rewards are utilized to represent domain knowledge and guide the training of RL models using expert knowledge. In our context, we aggregate the rewards from malfunctioning intersections to each working agent, with different weights, representing influences between malfunctioning and working agents. The additional reward guides working agents in earning benefits for malfunctioning intersections.
We employ a diffusion process without trainable parameters to perform a weighted sum of rewards. 
 The formulation is as follows:
\begin{equation}
    R'=\left[\sum_{k=1}^{K}(D_O^{-1}W)^k \odot Mask \right]R
\end{equation}
where $R\in \mathbb{R}^N$ is the vector of rewards of all intersections, $k$ is the diffusion step, $D_O^{-1}W$ is the state transition matrix. 
Similar to the state aggregation method described above, for i-th agent, we add the global reward $R'_{i}$ and the local reward $R_{i}$ to obtain the final reward, denoted as $R''_{i}$:
\begin{equation}
    R''_{i}=R_{i}+R'_{i}, i\in \{1,2,3,...,N\}
\end{equation}

\subsection{Training} \label{training}
MalLight is updated using the Bellman Equation\cite{ding2020introduction}:
\begin{equation}
    Q(S_{i}^{''t},a_{i}^{t})=R_{i}^{''t}+\gamma maxQ(S_{i}^{''t+1},a_{i}^{t+1})
\end{equation}
In this equation, $R_{i}^{''t}$ represents the reward that i-th agent can obtain by taking action $a_{i}^{t}$ based on the state $S_{i}^{''t}$, at decision time $t$. $\gamma$ is discount factor. We employ the Mean Squared Error (MSE) as the loss function to measure the disparity between the current action-value estimation and the desired action-value estimation provided by the Bellman Equation. The RL model is optimized using the RMSprop\cite{hinton2012neural} algorithm.

\section{Experiments}
\noindent In this section, we conduct experiments to answer the following research questions:
\\\noindent$\bullet$~\textbf{\textit{RQ1}}: Does MalLight outperform other SoTA methods on the malfunctioning signal scenario?
\\\noindent$\bullet$~\textbf{\textit{RQ2}}: Does MalLight perform well on the normal traffic signal control scenario?
\\\noindent$\bullet$~\textbf{\textit{RQ3}}: How the components in MalLight contribute to the performance?
\\\noindent$\bullet$~\textbf{\textit{RQ4}}: Does MalLight decrease accidental risk?
\\\noindent$\bullet$~\textbf{\textit{RQ5}}: How the diffusion process distributes state and reward aggregation weights?
\\\noindent$\bullet$~\textbf{\textit{RQ6}}: How do hyper-parameters affect the performance?
\\\noindent$\bullet$~\textbf{\textit{RQ7}}: How does the number of malfunctioning intersections affect the performance?

\vspace{-5pt}
\subsection{Experiment Setting}
\vspace{-8pt}
\begin{table}[ht]
\centering
\caption{SUMO Environment Setting}
\vspace{-8pt}
\begin{tabular}{ccc}
\hline
Attribute&Value&Description\\
\hline
accel&2.0&max acceleration ability ($m/s^2$)\\
decel&4.5&max deceleration ability ($m/s^2$)\\
length&5.0&vehicle netto-length ($m$)\\
width&2.0&vehicle width ($m$)\\
maxSpeed&40&max velocity ($km/h$)\\
minGap&2.5&empty space after leader ($m$)\\
jmIgnoreFoeProb&0.05&the probability of vehicle ignoring foes\\
\hline
\end{tabular}
\label{table-vehicle}
\end{table}

\vspace{-8pt}
\noindent\textbf{Environment setting}:
The experiments were conducted on a server with two Intel Xeon E5-2650 processors, 252GB of main memory, and two GeForce GTX 2080ti GPUs.
We conducted experiments based on the LibSignal framework~\cite{mei2022libsignal}, which is a traffic signal simulation toolkit developed on top of simulator engines SUMO~\cite{SUMO2018} and CityFlow~\cite{zhang2019cityflow,da2024cityflower}. We used SUMO for our experiments due to its flexibility in simulating intersection malfunction.
The SUMO settings are provided in Table~\ref{table-vehicle}. We generated vehicle trajectories by running the Dijkstra algorithm on each origin-destination pair in our datasets (introduced later). 
After feeding the traffic flow data into the simulator, each vehicle moves toward its destination following the predefined trajectory.

To better simulate traffic congestion and potential traffic accidents, we employ the FoeIgnore strategy~\cite{du2023safelight}. 
This mechanism enables each vehicle to ignore vehicles already within the intersection and potentially collide with them based on a predetermined probability. By simulating collisions at intersections, it mimics the congestion and chaos resulting from signal malfunctions.

\noindent\textbf{Model Setting}: 
For the traffic control module, which makes decisions based on local and global information, we use a deep Q-learning model with a Multi-Layer Perceptron (MLP) model as the backbone. Details of MLP are shown in Table~\ref{table-NN}. The training algorithm is described in Section~\ref{training}. Our model is trained for 200 episodes with a replay buffer size of 5000. For each episode, we run SUMO over the training dataset, with the signal phase determined by the current model.
The model is updated with 10 iterations over all MDP samples in a replay buffer at the end of each episode.
The learning rate is set to 0.001.
The diffusion step $K$ is 10. Discount factor $\gamma$ is set to 0.95. 

\begin{table}[ht]
    \centering
    \caption{Details of Backbone MLP}
    \vspace{-8pt}
    \begin{tabular}{cc}
    \hline
    Layer&$\#$Neurons\\
    \hline
    Input Layer&12 \\
    Hidden Layer 1&20 \\
    Hidden Layer 2&20 \\
    Output Layer&8\\
    \hline
    \end{tabular}
    \label{table-NN}
\end{table}

\vspace{-8pt}
\subsection{Datasets}\label{section:datasets}
\begin{figure}[ht]
\centering
    \begin{subfigure}[b]{0.48\linewidth}
        \centering
        \includegraphics[height=3cm]{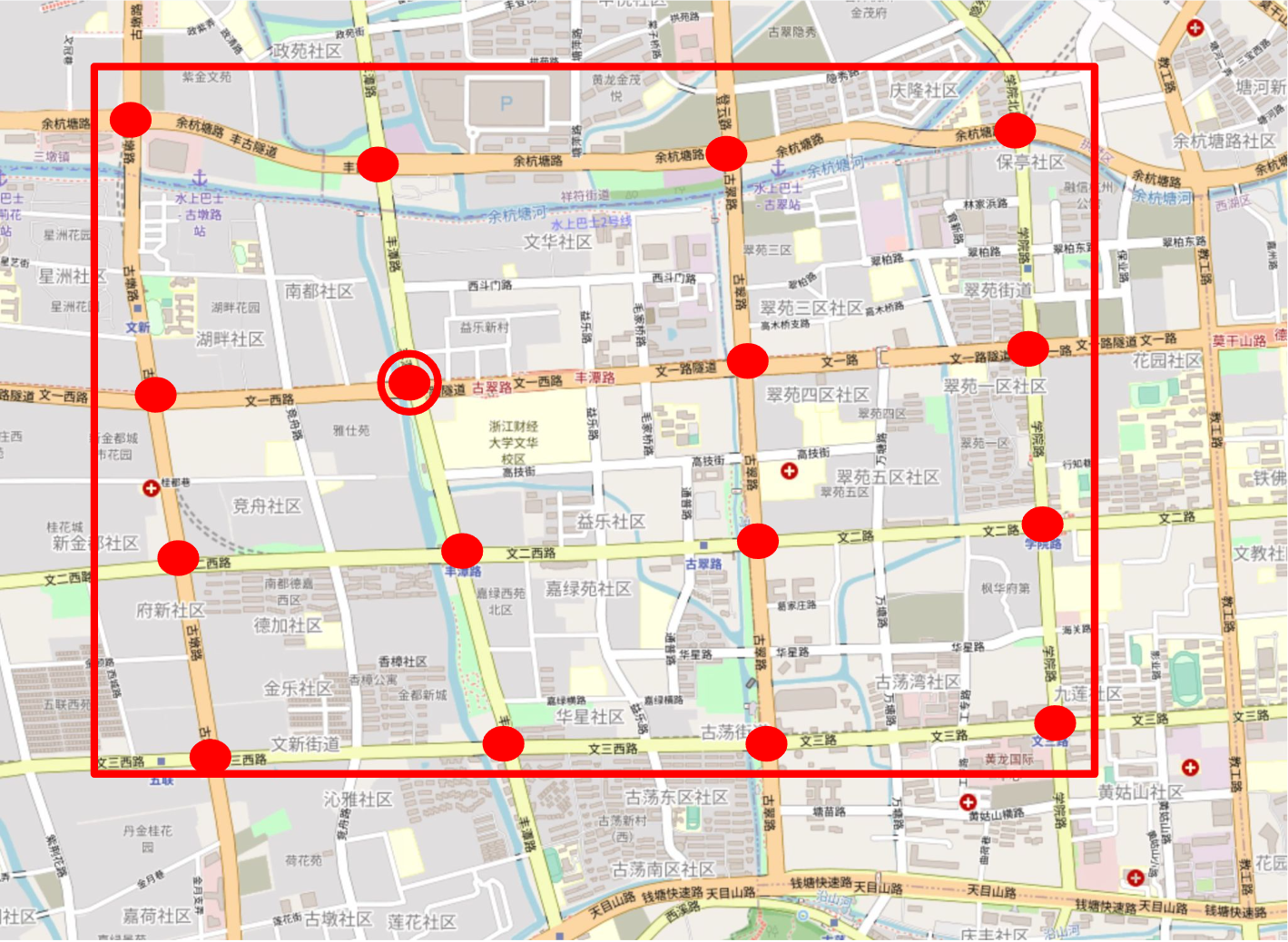}
        \caption{Gudang Sub-district, Hangzhou, China}
        \label{subfigure-Gudang}
    \end{subfigure}
    \hspace{0.1cm}
    \begin{subfigure}[b]{0.48\linewidth}
        \centering
        \includegraphics[height=3cm]{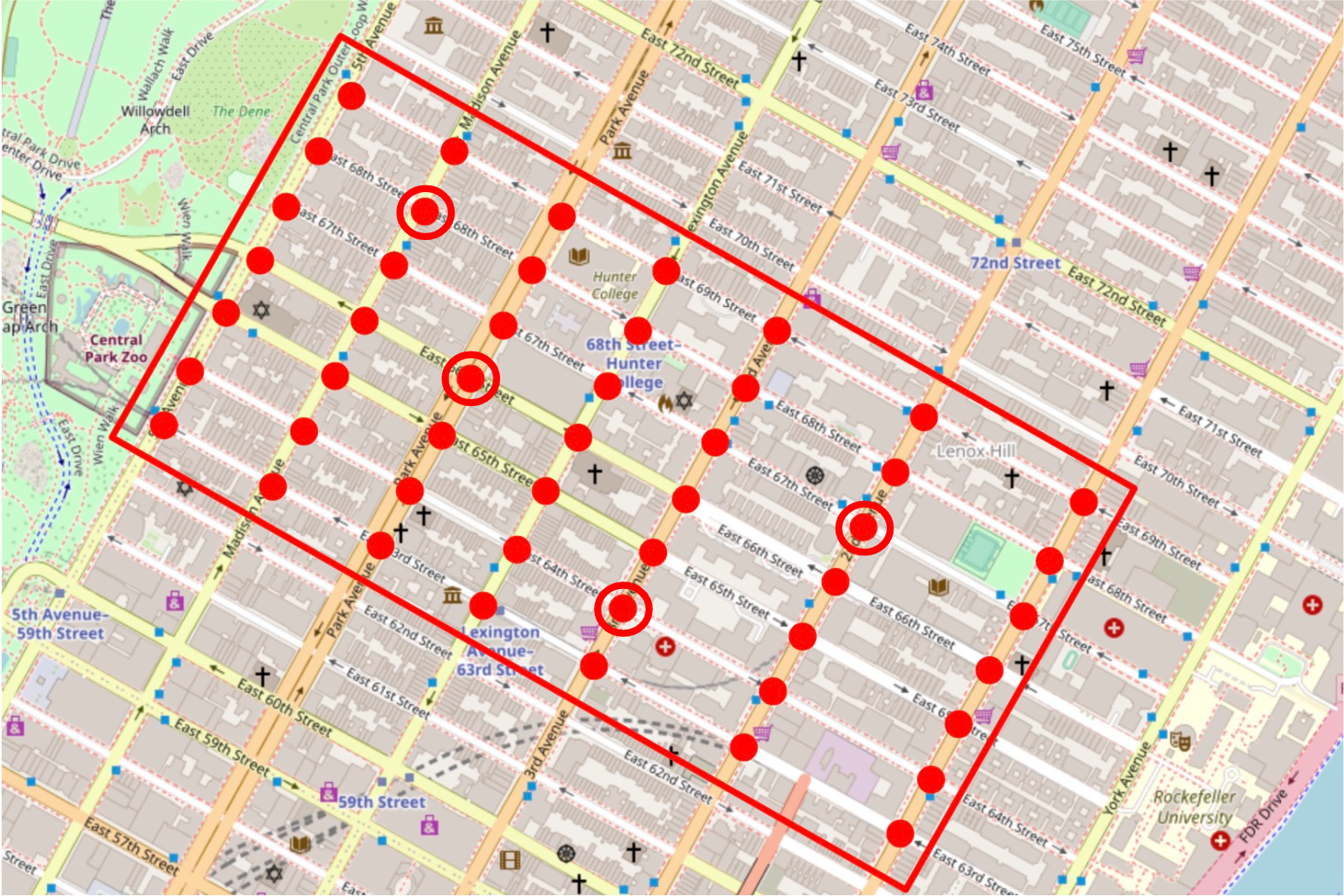}
        \caption{Upper East Side, Manhattan, New York, USA}
        \label{subfigure-NY}
    \end{subfigure}
\vspace{-8pt}
\setlength{\belowcaptionskip}{-20pt}
\caption{Road networks for real-world datasets. Red dots mark intersections chosen for experiments. Red circles mark malfunctioning signals.}
\label{figure-dataset}
\end{figure}

\noindent We use two real-world and one synthetic traffic flow data in our experiments. 
The road networks for real-world datasets were imported from OpenStreetMap~\cite{OpenStreetMap}, as shown in Figure~\ref{figure-dataset}.
The data statistics are listed in Table~\ref{table-dataset}. The road network of the synthetic Grid4*4 dataset is the same as that of Hangzhou dataset, and the only difference between them is the traffic flow pattern.
In datasets, a vehicle is described as $(o,d,t)$, where $o$ denotes the origin location, $d$ denotes the destination location, $t$ represents the departure time. Both origin ($o$) and destination ($d$) locations are within the road network.

\noindent$\bullet$~ \textbf{\textit{Hangzhou Dataset}}~\cite{datasets}: This is a widely used real-world dataset in prior works~\cite{wei2019colight,mei2023reinforcement,wu2021dynstgat}.
We selected 16 intersections in the Gudang Sub-district of Hangzhou and the signal at one of these intersections is set to be malfunctioning. Vehicle information was collected from roadside surveillance cameras. Each record in the camera data includes time, camera ID, and vehicle details.
The camera ID indicates the vehicle's location. Through the analysis of these records, we generated vehicle trajectories as they passed through the selected intersections. 

\noindent$\bullet$~\textbf{\textit{New York Dataset}}~\cite{NYC-Taxi-Trips}: This is also a well-known real-world dataset used in prior works~\cite{chen2020toward,wei2019presslight,wei2019colight,mei2023reinforcement}.
We take 49 intersections in the Upper East Side of Manhattan as a demonstration. 
In this setting, the traffic signals on four intersections are set to be malfunctioning.
Vehicle information was generated using open-source taxi trip data~\cite{NYC-Taxi-Trips}. This taxi trip data includes the geo-locations of the origin and destination for each trip. We initially mapped these geo-locations to the nearest intersections and then retained the trips that fell within the selected area. 
Because taxi trips can be seen as a sample from real-world trip
distributions, we empirically scaled up the number of trips by 4.5 to consider the impacts of personal and commercial vehicles. We made the simulation traffic volume roughly aligned with the real-world traffic volume indicated by New York Traffic Data Viewer~\cite{TrafficDataViewer}.

\noindent$\bullet$~\textbf{\textit{Grid4*4 Dataset}}: A synthetic dataset. The road network of Grid4*4 is the same as that of the Hangzhou dataset. We generate vehicles with a steady arrival rate and random origin and destination locations. Table ~\ref{table-dataset} shows details of generated traffic flow.

Each dataset comprises 2 hours of traffic flow data, of which we utilize 1 hour for training purposes and the remaining hour for testing.

\begin{table}[ht]
\centering
\caption{Data statistic of real-world traffic dataset}
\vspace{-8pt}
\begin{tabular}{cccccc}
\hline
\multirow{2}*{Dataset}&\multirow{2}*{$\#$intersections}&\multicolumn{4}{c}{Arrival rate ($\#$vehicles/300s)}\\
&&Mean&Std&Max&Min\\
\hline
Hangzhou&16&1053.26&86.70&1352&512\\
New York&49&361.19&10.08&411&324\\
Grid4*4&16&1200&0&1200&1200\\
\hline
\end{tabular}
\label{table-dataset}
\end{table}

\newcolumntype{Y}{>{\centering\arraybackslash}X}
\begin{table*}[t]
\centering
\caption{Results on Hangzhou, New York and Grid4*4 datasets}
\vspace{-8pt}
\begin{tabularx}{0.9\textwidth}{|c|Y|Y|Y|Y|Y|Y|Y|}
\hline
\multirow{2}*{Methods}&\multicolumn{4}{c|}{Intersection throughput}&\multicolumn{3}{c|}{Road network throughput}\\
\cline{2-8}
&No Mal $\uparrow$&Mal $\uparrow$&RR ($\%$) $\downarrow$&$\#$Acc $\downarrow$&No Mal $\uparrow$&Mal $\uparrow$&RR ($\%$) $\downarrow$\\
\hline
\multicolumn{8}{|c|}{\textbf{\textit{Hangzhou}}}\\
\hline
FixedTime&538&390&27.5&18&2923&2653&9.2\\
SOTL&760&489&35.7&23&3674&3118&15.1\\
MaxPressure&592&405&31.5&27&3997&3777&5.5\\
IDQN&676&387&42.8&36&4554&4133&9.2\\
CoLight&702&457&34.9&31&4617&4219&8.6\\
Advanced-CoLight&740&526&28.9&27&4679&4295&8.2\\
Advanced-PressLight&752&561&25.3&25&4745&4370&7.9\\
PRLight&\underline{\textbf{841}}&577&31.4&29&\underline{\textbf{4779}}&4310&9.8\\
\textbf{MalLight} (ours)&822&\underline{\textbf{715}}&\underline{\textbf{13}}&\underline{\textbf{11}}&4607&\underline{\textbf{4416}}&\underline{\textbf{4.2}}\\
\hline
\hline
\multicolumn{8}{|c|}{\textbf{\textit{New York}}}\\
\hline
FixedTime&540&365&32&31&2671&2239&16.1\\
SOTL&579&492&15&18&3172&2594&18.2\\
MaxPressure&630&570&9&13&3068&2852&7\\
IDQN&617&554&10.2&22&3223&2883&10.5\\
CoLight&627&561&10.5&18&3507&3119&11.1\\
Advanced-CoLight&625&562&10.1&16&3612&3154&12.6\\
Advanced-PressLight&\underline{\textbf{642}}&542&15.5&17&\underline{\textbf{3784}}&3276&13.4\\
PRLight&639&534&16.4&19&3695&3117&15.6\\
\textbf{MalLight} (ours)&633&\underline{\textbf{592}}&\underline{\textbf{6.5}}&\underline{\textbf{8}}&3532&\underline{\textbf{3309}}&\underline{\textbf{6.3}}\\
\hline
\hline
\multicolumn{8}{|c|}{\textbf{\textit{Grid4*4}}}\\
\hline
FixedTime                 &579&381&34.2&25&3192&2582&19.1 \\
SOTL                      &612&418&31.7&23&3463&2829&18.3 \\
MaxPressure               &645&454&29.6&21&3728&3053&18.1 \\
IDQN                      &796&504&36.7&29&4339&3454&20.4 \\
CoLight                   &832&589&29.2&19&4591&3778&17.7 \\
Advanced-CoLight          &854&624&26.9&17&4829&4076&15.6 \\
Advanced-PressLight       &883&679&23.1&17&4914&4236&13.8 \\
PRLight                   &\underline{\textbf{892}}&684&23.3&15&\underline{\textbf{4963}}&4258&14.2 \\
\textbf{MalLight}(ours) &877&\underline{\textbf{767}}&\underline{\textbf{12.5}}&\underline{\textbf{9}}&4871&\underline{\textbf{4515}}&\underline{\textbf{7.3}} \\
\hline

\multicolumn{8}{l}{No Mal denotes No Malfunction, Mal denotes Malfunction, RR denotes Reduction Ratio, $\#$Acc denotes $\#$Accidents}
\end{tabularx}
\label{table-result}
\end{table*}
\vspace{-10pt}

\subsection{Evaluation Metrics}
Following existing studies~\cite{wei2019survey}, we mainly use \textit{throughput} to evaluate the performance of different models for traffic signal control.
In addition, we simulate the number of accidents to evaluate the impacts of our model on reducing accidental risk.
The detailed descriptions are as follows:
\\\noindent$\bullet$~\textbf{\textit{Network Throughput}}: It represents the number of vehicles that have completed their trip in the road network during the one-hour testing period.
\\\noindent$\bullet$~\textbf{\textit{Intersection Throughput}}: It represents the (average) number of vehicles that pass through malfunctioning intersection(s) during the one-hour testing period.
We introduced this intersection-level metric to better compare the performance of models on malfunctioning intersections.
\\\noindent$\bullet$~\textbf{\textit{Number of Accidents}}: It represents the number of traffic collisions simulated by SUMO in malfunctioning intersections. This is an additional metric. We do not calculate collisions at the road network level considering traffic collisions rarely occur in signalized intersections on SUMO.

\vspace{-8pt}
\subsection{Compared Methods}\label{section:compared methods}
\noindent$\bullet$~\textbf{\textit{FixedTime}}~\cite{koonce2008traffic}: A policy with predefined fixed phase length and phase splits, which is commonly used in practice. The offset is randomly set in our experiments.
\\\noindent$\bullet$~\textbf{\textit{SOTL}}~\cite{cools2013self}: A rule-based method based on the request from current phase and competing phases.
\\\noindent$\bullet$~\textbf{\textit{MaxPressure}}~\cite{varaiya2013max}: A state-of-the-art (SoTA) optimization-based algorithm that minimizes pressure at each intersection. 
It reduces over-saturation by balancing queue lengths between each pair of neighboring intersections.
\\\noindent$\bullet$~\textbf{\textit{IDQN}}~\cite{wei2018intellilight}: A deep RL method in which information is not shared among agents. Each intersection is controlled by one agent, and these agents do not share model parameters.
\\\noindent$\bullet$~\textbf{\textit{CoLight}}~\cite{wei2019colight}: A deep RL method based on Graph Attention Network. 
\\\noindent$\bullet$~\textbf{\textit{Advanced-CoLight}}~\cite{zhang2022expression}: A SoTA RL algorithm template that incorporates advanced traffic state (ATS) into the RL model. It uses CoLight as a backbone network.
\\\noindent$\bullet$~\textbf{\textit{Advanced-PressLight}}~\cite{zhang2022expression}: Similar to Advanced-CoLight, it uses PressLight~\cite{wei2019presslight} as the backbone network. PressLight is a deep RL method that coordinates traffic signals by minimizing the Pressure~\cite{varaiya2013max} at each intersection to maximize the throughput of the road network.
\\\noindent$\bullet$~\textbf{\textit{PRLight}}~\cite{han2023mitigating}: A SoTA RL method that integrates a dynamic graph representation module and a traffic prediction module to effectively control traffic signals and mitigate action hysteresis in real-time traffic management.

\vspace{-12pt}
\subsection{Experiment Result}
\subsubsection{Overall Performance}
Our experimental results using Hangzhou dataset and New York dataset are shown in Table~\ref{table-result}. 
By analyzing the results, we have the following findings:
\\\noindent$\bullet$~\textbf{\textit{RQ1}}: 
By analyzing the Reduction Ratio (RR) columns in Table~\ref{table-result}, we can conclude that our model, MalLight, performs better in malfunctioning scenarios. When a malfunction occurs, the throughput reduction ratios, both at the intersection level and network level, are lower than those of other models. This suggests that our proposed state and reward modules genuinely enhance the resilience of the RL model to intersection malfunctions. By reducing the traffic demand in malfunctioning intersections, congestion is alleviated, and malfunctioning intersections experience a lower throughput reduction. Specifically, on Hangzhou dataset, compared with the second-best model Advanced-PressLight, our model alleviates throughput reduction by as much as 48.6$\%$.
\\\noindent$\bullet$~\textbf{\textit{RQ2}}:
By analyzing the throughput when no malfunction, we conclude that MalLight performs well in normal (without malfunction) scenarios. The throughputs at both the intersection and network level when no malfunction are very close to the highest throughputs of other models. In addition, traditional Methods (FixedTime, SOTL, MaxPressure) perform poorly in both intersection and network levels. This is because traffic flow in our datasets is highly dynamic. These traditional methods rely heavily on human-defined assumptions of the environment, which may not be true under some circumstances.
\\\noindent$\bullet$~An interesting finding is that the throughput reduction of the whole road network can be greatly higher than the throughput reduction of malfunctioning intersections, which means that congestion in malfunctioning intersections affects other well-functioning intersections. 
This occurs because when congestion arises in malfunctioning intersections, waiting vehicles can spill over into neighboring intersections, leading to congestion in those neighboring intersections and a subsequent reduction in their throughput. 
For example, in the case of SOTL under the New York dataset, the total throughput reduction of malfunctioning intersections is $348$, while the throughput reduction for the entire road network is 578, which is 66.1\% higher than that of the malfunctioning intersections.

\subsubsection{Ablation Study}
We consider several variations of MalLight:
\\\noindent$\bullet$~\textbf{\textit{MalLight-S}}: For state aggregation, instead of using diffusion convolution which contains trainable parameters, MalLight-S uses a diffusion process without any trainable parameter.
This variant is to show the benefits of the flexible state aggregation in our model.
\\\noindent$\bullet$~\textbf{\textit{MalLight-R}}: We remove reward aggregation in MalLight-R.
This means that each agent exclusively concentrates on its own reward and disregards the benefits of malfunctioning intersections.
\\\noindent$\bullet$~\textbf{\textit{MalLight-M}} We remove Malfunction Mask from MalLight. MalLight-M allows each agent to observe and care about the entire road network.

\vspace{-6pt}
\begin{table}[h]
\centering
\caption{Results of Ablation Study}
\vspace{-8pt}
\begin{tabular}{|c|c|c|c|}
\hline
\multirow{3}*{Methods}&\multicolumn{3}{c|}{Intersection Throughput Reduction $(\%)$ $\downarrow$}\\
\cline{2-4}
&Hangzhou&New York&Grid4*4\\
\hline
\textbf{MalLight*}&\underline{\textbf{13}}&\underline{\textbf{6.5}}&\underline{\textbf{12.5}}\\
MalLight-S&18.8&9.9&20.3\\
MalLight-R&30.9&16.1&27.4\\
MalLight-M&26.3&11.3&23.6\\
\hline
\end{tabular}
\label{table-result-ablation}
\end{table}

Table~\ref{table-result-ablation} shows the performance of variations of our model. We have the following findings (\textbf{\textit{RQ3}}):
\\\noindent$\bullet$~The trainable parameters in diffusion convolution provide the model with greater adaptability, thereby enhancing its performance.
\\\noindent$\bullet$~The absence of reward aggregation, which makes the agent prioritize caring about malfunctioning intersections, leads to a significant decrease in performance.
\\\noindent$\bullet$~MalLight outperforms MalLight-M due to the presence of the Malfunction Mask, which filters out well-functioning intersections and enables each agent to concentrate on what we want it to pay attention to—malfunctioning intersections. Features and rewards coming from well-functioning intersections can be distracting for the agent, under which circumstances the agent may struggle to focus on malfunctioning intersections or even itself.

\subsubsection{Accident Risk (\textbf{\textit{RQ4}})}:
We analyze the relationship between accident risk and throughput reduction. The number of accidents serves as an additional metric for our model. When an accident occurs at an intersection, the victims can block following vehicles, leading to congestion. By examining the column labeled $\#$Acc, we can observe that a higher level of Reduction Ratio typically corresponds to a higher number of accidents.

\subsubsection{Weight Distribution from Diffusion Process (\textbf{\textit{RQ5}})}
\begin{figure}[ht]
\centering
\setlength{\belowcaptionskip}{-15pt}
    \begin{minipage}[t]{0.48\linewidth}
        \centering
        \includegraphics[width=4.2cm]{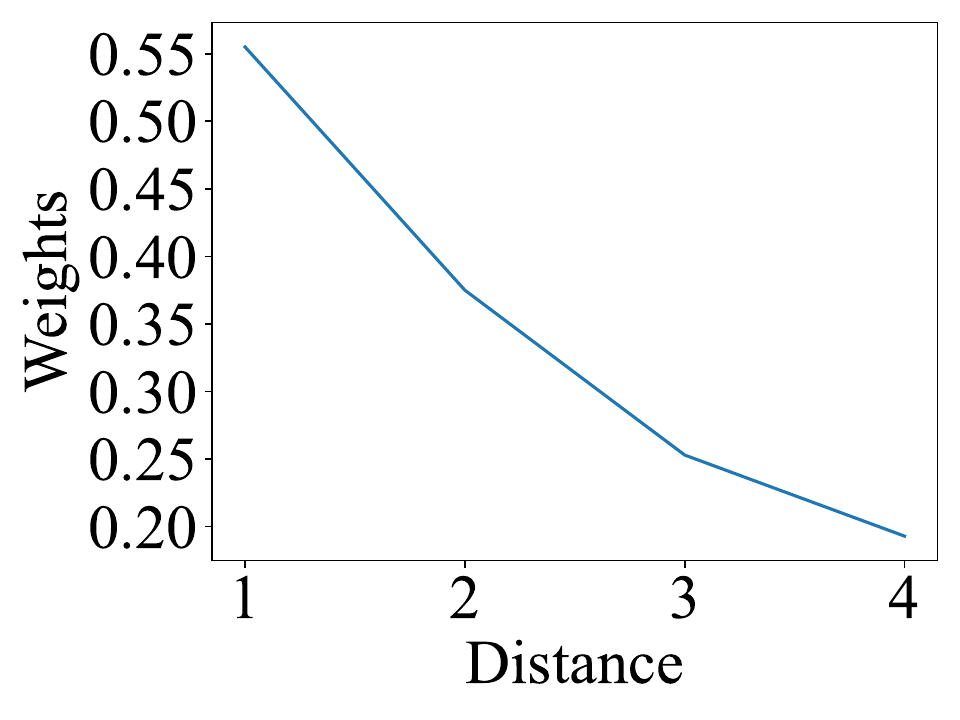}
        \vspace{-20pt}
        \caption{Impacts of intersection distances on their influence}
        \label{fig:figure-DPWeights}
    \end{minipage}
    \hspace{0.1cm}
    \begin{minipage}[t]{0.48\linewidth}
        \centering
        \includegraphics[width=4.2cm]{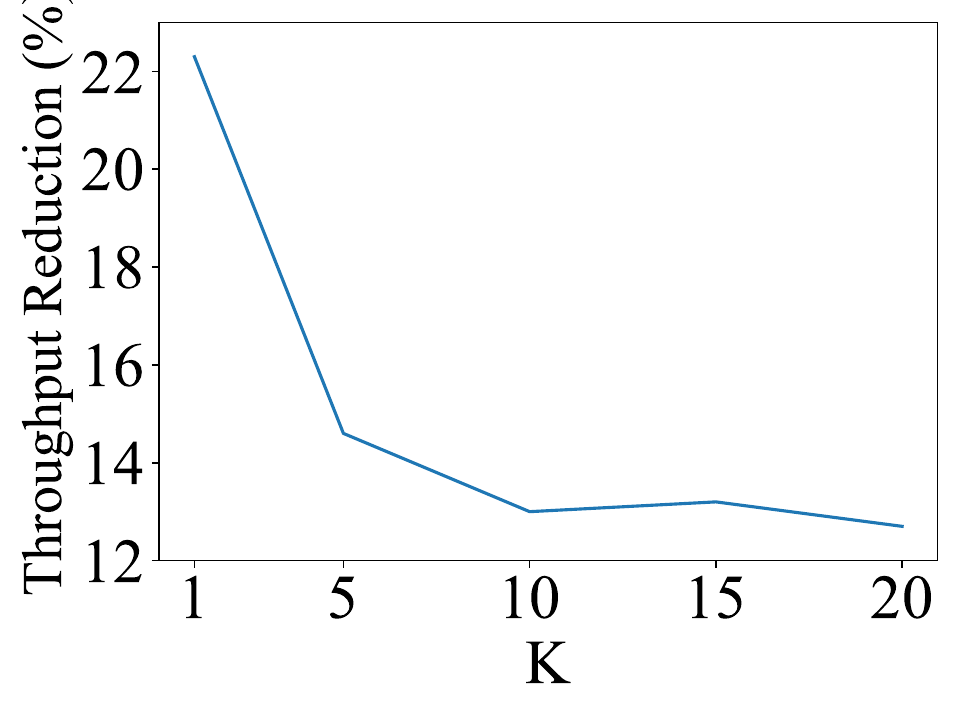}
        \vspace{-20pt}
        \caption{Impacts of diffusion steps K on throughput reduction ratio}
        \label{fig:figure-K}
    \end{minipage}
\end{figure}

We select the upper-left intersection in Hangzhou road network as an example and calculate the influence weights generated by a diffusion process. 
Figure~\ref{fig:figure-DPWeights} demonstrates the influence (weights) from this node to other nodes at distances of 1 to 4 hops away. 
Because the road network is a bidirectional graph, with two edges connecting a pair of nodes and sharing the same initial weight (determined by road distance processed through a thresholded Gaussian kernel), this figure can also be viewed as representing the influence from this specific node to other nodes at varying distances (hops) in the graph. We can observe that as the distance (number of hops in the graph) increases, the influence between two nodes tends to decrease.

\subsubsection{Parameter Sensitivity (\textbf{\textit{RQ6}})}
We set different $K$ values and observe the throughput reduction metric. 
The parameter $K$ in the diffusion process, as well as its counterpart in diffusion convolution, determines how many hops a signal can diffuse along edges.
For example, if $K=1$, the malfunctioning intersection will only consider directly connected intersections.
The result is shown in Figure~\ref{fig:figure-K}.
When $K$ is set too low, the observation scope of well-functioning intersections will be significantly limited, and the influence among intersections cannot be adequately modeled. It's worth noting that a higher value of $K$ requires more computational resources.
Based on the curve, we set $K=10$ in experiments.

\subsubsection{Performance on various number of malfunctioning intersections (\textbf{\textit{RQ7}})}
\begin{figure}[ht]
    \centering
    \includegraphics[width=4.2cm]{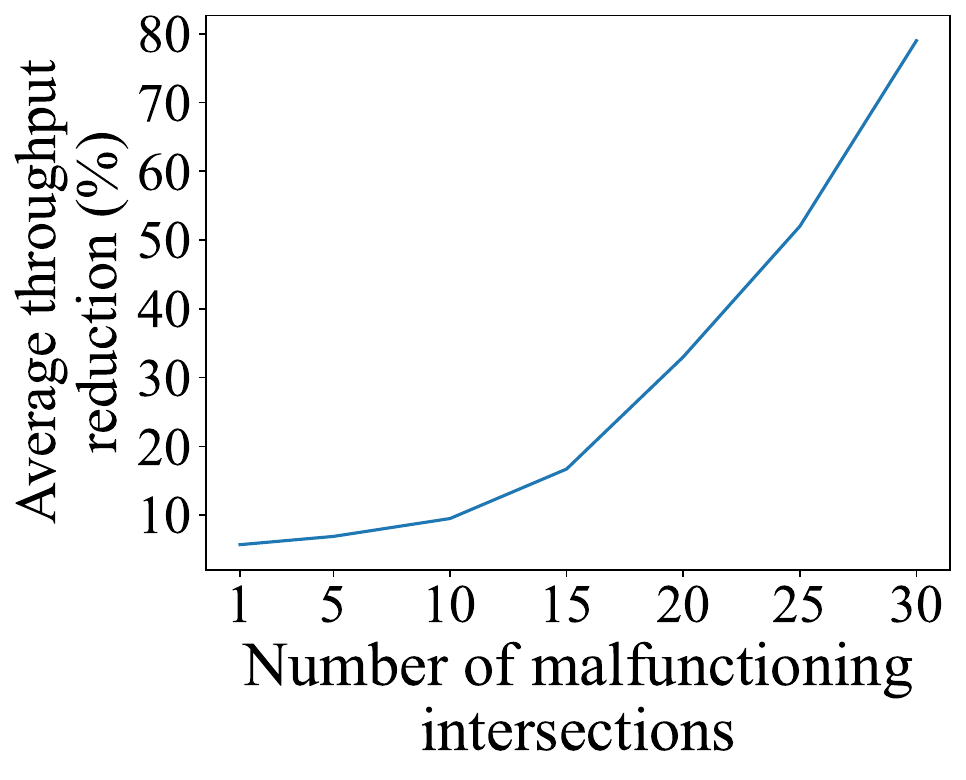}
    \vspace{-8pt}
    \caption{Performance on different $\#$malfunctioning intersections}
    \label{fig:variousIntersections}
\end{figure}

We conduct experiments using the MalLight framework on the New York dataset, varying the number of malfunctioning signals. For each experiment, our model is run five times, and scores are averaged. The result is shown in Figure~\ref{fig:variousIntersections}. As the number of malfunctioning signals increases, the curve initially remains relatively flat before experiencing a sharp increase after approximately 15 signals.
Our model operates under the assumption that nearby working intersections can provide assistance to malfunctioning ones. Consequently, if the percentage of malfunctioning intersections exceeds a certain threshold, the effectiveness of the model decreases significantly.

\vspace{-18pt}
\section{Discussion}
\noindent\textbf{Lessons Learned}:
We summarize key lessons learned from work:
\\\noindent$\bullet$ We design our malfunction-resistant approach based on inter-intersection cooperation, as manifested in the design of state and reward modules. In contrast to alternative methods that neglect the advantageous potential of well-functioning intersections in alleviating the plight of malfunctioning counterparts, our approach stands out by affording a notably reduced reduction in throughput.
\\\noindent$\bullet$ The integration of state and reward aggregation mechanisms, while taking into consideration the influences among agents, is of great importance in coordinating intersections and enhancing the model's resilience in the face of signal malfunction.
In direct comparison to the Advanced-PressLight model which shares similar architecture but lacks the incorporation of these two pivotal modules, our approach achieves better performance in the presence of signal malfunction.
In essence, by optimizing control of surrounding intersections, the impact of signal malfunction can be alleviated, thus aligning with our central research idea.

\noindent\textbf{Limitations}:
Constrained by the inherent limitations of the driver behavior model within SUMO, regrettably fails to authentically replicate the congestion resulting from signal malfunctions, we resorted to the use of traffic collisions as a surrogate for congestion induction. We have to acknowledge that not all instances of congestion are precipitated by traffic collisions. In the future, our objective is to enhance the fidelity of our simulation by refining the driver behavior model to more accurately emulate real-world congestion dynamics.

\vspace{-10pt}
\section{Related Work}
\subsection{Traffic Signal Malfunction}
The time series of traffic signal malfunctions and the subsequent actions are as follows: malfunction occurs, detection, and response.
Methods designed to address malfunctioning traffic signals can be categorized into two groups: detection and response methods.

\textbf{\textit{Detection methods}}: As stated in~\cite{chen2009assessment}, malfunction detection typically relies on manual reports.
Several resources~\cite{what-should-you-do-if-a-traffic-is-malfunctioning} are available for reporting signal problems, including 24-hour hotlines and 911 calls. 
Manual detection, based on road user reports, often results in delays. 
Remote monitoring techniques may be employed in some areas when crew time and workload allow, but they are seldom effective due to the expensive human resources required.
Furthermore, some automatic detection systems have been developed. 
Purba et al~\cite{purba2018developing} develop a self-diagnosis system to detect traffic signal malfunction immediately. 
Soh et al~\cite{soh2013smart} design a monitoring system based on fuzzy technology for malfunction detection.
In our work, we assume malfunctioning signals are detected by existing work.

\textbf{\textit{Response methods}}: 
The response to put malfunctioning traffic signals back to fully functional relies heavily on manual repair or replacement actions taken by contractors or authorities~\cite{chen2009assessment}.
In addition to manual response, some automatic response methods are also designed. 
Malfunction flash~\cite{benioff1980study} is designed to prevent safety issues from traffic signal malfunction. 
When an error is detected, the signal is automatically placed into flash mode as a safety precaution (if the signal is still accessible). However, malfunction flash only serves as a warning to road users. 
However, some decisions, such as whether to cross or wait, and whether to stop before crossing, are left to drivers, based on their observation of traffic conditions. 
Hunter et al~\cite{hunter2011mode} studied the driver behaviors under malfunction flash and revealed that malfunction flash cannot eliminate the safety issue of traffic signal malfunctions, not to mention other impacts such as congestion. In addition, our method is applicable in situations where the traffic signal is in a blackout, where the traffic signal cannot even work as a flashing warning to drivers.

Some methods are designed for efficient human resources distribution. For example, Mathibela~\cite{mathibela2017another} designs a RUSBoost-based framework to predict the best distribution of human resources at critical intersections in the presence of malfunctioning traffic lights.
Our work is considered an immediate response before human resources get involved to prevent the heavy consequences resulting from traffic signal malfunction.

\subsection{Traffic Signal Control}
Traffic Signal Control is an important sub-domain of smart city~\cite{sc1,sc2,sc3,sc4,sc5,sc6,sc7,sc8,sc9}.
Methods for controlling traffic signals can be divided into classic optimization-based methods~\cite{varaiya2013max,cools2013self,stevanovic2009scoot} and RL-based methods~\cite{wei2018intellilight,wei2019presslight,zhang2022expression,wei2019colight,han2023mitigating}. 
Among these techniques, RL is most popular these years because of its ability to learn directly from complex conditions without assumptions about the environment~\cite{rl1,rl2,rl3,vlachogiannis2023humanlight,jiang2024x,ruan2024coslight,jiang2024guidelight,da2024open}. Thus, RL-based methods and Deep Learning based methods usually outperform classic optimization-based methods.
However, current RL-based methods exhibit suboptimal performance under scenarios involving signal malfunctions due to the isolation of reward mechanisms. Specifically, each agent operates with a focus exclusively on optimizing the traffic flow at its own intersection, neglecting the potential benefits of adjacent intersections. While this approach is effective in standard conditions, where each intersection is regulated by a functioning agent, it falters in the event of a signal malfunction. In such cases, the affected intersection lacks the capability to leverage support from nearby intersections, causing increased congestion.

Recently, a number of works have focused on traffic disruptions, such as missing data~\cite{mei2023reinforcement}, incidents~\cite{du2023safelight,rodrigues2019towards}, or weather changes~\cite{da2024prompt,da2023sim2real,da2023uncertainty}. 
Traffic signal malfunction is different from these disruptions in that: Firstly, these disruptions affect the RL model's input, while signal malfunction disables the RL agent from interacting with the environment, i.e. affects the output. Secondly, these disruptions test a single model's robustness, while signal malfunction tests the coordination ability of the whole system. Because a malfunctioning signal loses the ability to control traffic flow, it is naturally in need of help from other well-functioning signals, which makes cooperation an important factor in this context. In conclusion, traffic signal malfunction is a unique scenario that has not been considered in previous traffic disruption and robustness works.

\vspace{-5pt}
\section{Conclusion}
In this paper, we study the problem of intelligent traffic signal control in the presence of traffic signal malfunction. We design MalLight, an traffic signal control model founded on RL principles.
In the MalLight framework, we introduce two pioneering components, namely the Influence-aware State Aggregation Module and the Influence-aware Reward Aggregation Module, which are seamlessly integrated into the RL architecture. We subsequently conduct a comprehensive array of empirical investigations on two real-world datasets. Our empirical findings demonstrate that MalLight surpasses the performance of extant methods in both normal conditions and scenarios characterized by signal malfunctions. Notably, our results reveal a substantial improvement in the reduction of intersection throughput, with a remarkable mitigation rate of up to 48.6$\%$.

\vspace{-8pt}
\section{Acknowledgments}
The work was supported in part by the National Science Foundation under Grant No. 2047822, 1952096, 2318697 and 2421839. We thank all the reviewers for their insightful feedback to improve this paper.

\bibliographystyle{ACM-Reference-Format}
\bibliography{ref}


\begin{thebibliography}{73}


\ifx \showCODEN    \undefined \def \showCODEN     #1{\unskip}     \fi
\ifx \showDOI      \undefined \def \showDOI       #1{#1}\fi
\ifx \showISBNx    \undefined \def \showISBNx     #1{\unskip}     \fi
\ifx \showISBNxiii \undefined \def \showISBNxiii  #1{\unskip}     \fi
\ifx \showISSN     \undefined \def \showISSN      #1{\unskip}     \fi
\ifx \showLCCN     \undefined \def \showLCCN      #1{\unskip}     \fi
\ifx \shownote     \undefined \def \shownote      #1{#1}          \fi
\ifx \showarticletitle \undefined \def \showarticletitle #1{#1}   \fi
\ifx \showURL      \undefined \def \showURL       {\relax}        \fi
\providecommand\bibfield[2]{#2}
\providecommand\bibinfo[2]{#2}
\providecommand\natexlab[1]{#1}
\providecommand\showeprint[2][]{arXiv:#2}

\bibitem[new(2023)]%
        {newarkconnect}
 \bibinfo{year}{2023}\natexlab{}.
\newblock \bibinfo{booktitle}{\emph{Newark Connect}}.
\newblock
\urldef\tempurl%
\url{https://www.newarknj.gov/card/newark-connect}
\showURL{%
\tempurl}
\newblock
\shownote{Accessed: 2023-06-30}.


\bibitem[NYC(2023)]%
        {NYC-Taxi-Trips}
 \bibinfo{year}{2023}\natexlab{}.
\newblock \bibinfo{booktitle}{\emph{NYC-Taxi-Trips}}.
\newblock
\urldef\tempurl%
\url{https://www.nyc.gov/site/tlc/about/tlc-trip-record-data.page/}
\showURL{%
\tempurl}
\newblock
\shownote{Accessed: 2023-06-30}.


\bibitem[Ope(2023)]%
        {OpenStreetMap}
 \bibinfo{year}{2023}\natexlab{}.
\newblock \bibinfo{booktitle}{\emph{OpenStreetMap}}.
\newblock
\urldef\tempurl%
\url{https://www.openstreetmap.org/}
\showURL{%
\tempurl}
\newblock
\shownote{Accessed: 2023-06-30}.


\bibitem[chi(2023)]%
        {chicago}
 \bibinfo{year}{2023}\natexlab{}.
\newblock \bibinfo{booktitle}{\emph{Some Chicago traffic lights still broken months, even a year, after reporting to 311}}.
\newblock
\urldef\tempurl%
\url{https://abc7chicago.com/chicago-traffic-light-broken-311-cdot/12300361/}
\showURL{%
\tempurl}
\newblock
\shownote{Accessed: 2023-06-30}.


\bibitem[Tra(2023)]%
        {TrafficDataViewer}
 \bibinfo{year}{2023}\natexlab{}.
\newblock \bibinfo{booktitle}{\emph{Traffic Data Viewer}}.
\newblock
\urldef\tempurl%
\url{https://www.dot.ny.gov/tdv/}
\showURL{%
\tempurl}
\newblock
\shownote{Accessed: 2023-06-30}.


\bibitem[dat(2023)]%
        {datasets}
 \bibinfo{year}{2023}\natexlab{}.
\newblock \bibinfo{booktitle}{\emph{Traffic Signal Control Datasets}}.
\newblock
\urldef\tempurl%
\url{https://traffic-signal-control.github.io/index.html\#open-datasets/}
\showURL{%
\tempurl}
\newblock
\shownote{Accessed: 2023-06-30}.


\bibitem[wha(2023)]%
        {what-should-you-do-if-a-traffic-is-malfunctioning}
 \bibinfo{year}{2023}\natexlab{}.
\newblock \bibinfo{booktitle}{\emph{What should you do if a traffic light is malfunctioning}}.
\newblock
\urldef\tempurl%
\url{https://elteccorp.com/news/other/what-should-you-do-if-a-traffic-light-is-malfunctioning/}
\showURL{%
\tempurl}
\newblock
\shownote{Accessed: 2023-06-30}.


\bibitem[Aftabuzzaman(2007)]%
        {aftabuzzaman2007measuring}
\bibfield{author}{\bibinfo{person}{Md Aftabuzzaman}.} \bibinfo{year}{2007}\natexlab{}.
\newblock \showarticletitle{Measuring traffic congestion-a critical review}. In \bibinfo{booktitle}{\emph{30th Australasian transport research forum}}. ETM GROUP London, UK, \bibinfo{pages}{1--16}.
\newblock


\bibitem[Al-Madani(2003)]%
        {al2003dynamic}
\bibfield{author}{\bibinfo{person}{Hashim~MN Al-Madani}.} \bibinfo{year}{2003}\natexlab{}.
\newblock \showarticletitle{Dynamic vehicular delay comparison between a police-controlled roundabout and a traffic signal}.
\newblock \bibinfo{journal}{\emph{Transportation Research Part A: Policy and Practice}} \bibinfo{volume}{37}, \bibinfo{number}{8} (\bibinfo{year}{2003}), \bibinfo{pages}{681--688}.
\newblock


\bibitem[Benioff et~al\mbox{.}(1980)]%
        {benioff1980study}
\bibfield{author}{\bibinfo{person}{B Benioff}, \bibinfo{person}{FC Dock}, {and} \bibinfo{person}{C Carson}.} \bibinfo{year}{1980}\natexlab{}.
\newblock \bibinfo{booktitle}{\emph{A STUDY OF CLEARANCE INTERVALS, FLASHING OPERATION, AND LEFT-TURN PHASING AT TRAFFIC SIGNALS. VOLUME 2. CLEARANCE INTERVALS}}.
\newblock \bibinfo{type}{{T}echnical {R}eport}.
\newblock


\bibitem[Cai et~al\mbox{.}(2023)]%
        {sc9}
\bibfield{author}{\bibinfo{person}{Kunlin Cai}, \bibinfo{person}{Jinghuai Zhang}, \bibinfo{person}{Will Shand}, \bibinfo{person}{Zhiqing Hong}, \bibinfo{person}{Guang Wang}, \bibinfo{person}{Desheng Zhang}, \bibinfo{person}{Jianfeng Chi}, {and} \bibinfo{person}{Yuan Tian}.} \bibinfo{year}{2023}\natexlab{}.
\newblock \showarticletitle{Where have you been? A Study of Privacy Risk for Point-of-Interest Recommendation}.
\newblock \bibinfo{journal}{\emph{arXiv preprint arXiv:2310.18606}} (\bibinfo{year}{2023}).
\newblock


\bibitem[Chen et~al\mbox{.}(2020)]%
        {chen2020toward}
\bibfield{author}{\bibinfo{person}{Chacha Chen}, \bibinfo{person}{Hua Wei}, \bibinfo{person}{Nan Xu}, \bibinfo{person}{Guanjie Zheng}, \bibinfo{person}{Ming Yang}, \bibinfo{person}{Yuanhao Xiong}, \bibinfo{person}{Kai Xu}, {and} \bibinfo{person}{Zhenhui Li}.} \bibinfo{year}{2020}\natexlab{}.
\newblock \showarticletitle{Toward a thousand lights: Decentralized deep reinforcement learning for large-scale traffic signal control}. In \bibinfo{booktitle}{\emph{Proceedings of the AAAI Conference on Artificial Intelligence}}, Vol.~\bibinfo{volume}{34}. \bibinfo{pages}{3414--3421}.
\newblock


\bibitem[Chen et~al\mbox{.}(2009)]%
        {chen2009assessment}
\bibfield{author}{\bibinfo{person}{Wenling Chen}, \bibinfo{person}{Larrie Henley}, {and} \bibinfo{person}{Jeff Price}.} \bibinfo{year}{2009}\natexlab{}.
\newblock \showarticletitle{Assessment of traffic signal maintenance and operations needs at Virginia department of transportation}.
\newblock \bibinfo{journal}{\emph{Transportation research record}} \bibinfo{volume}{2128}, \bibinfo{number}{1} (\bibinfo{year}{2009}), \bibinfo{pages}{11--19}.
\newblock


\bibitem[Chowdhury(2013)]%
        {chowdhury2013traffic}
\bibfield{author}{\bibinfo{person}{Md~Mamun Chowdhury}.} \bibinfo{year}{2013}\natexlab{}.
\newblock \showarticletitle{Traffic congestion and mismanagement in dhaka city}.
\newblock \bibinfo{journal}{\emph{Planned Decentralization: Aspired Development, World Town Planning Day}} (\bibinfo{year}{2013}).
\newblock


\bibitem[Cools et~al\mbox{.}(2013)]%
        {cools2013self}
\bibfield{author}{\bibinfo{person}{Seung-Bae Cools}, \bibinfo{person}{Carlos Gershenson}, {and} \bibinfo{person}{Bart D’Hooghe}.} \bibinfo{year}{2013}\natexlab{}.
\newblock \showarticletitle{Self-organizing traffic lights: A realistic simulation}.
\newblock \bibinfo{journal}{\emph{Advances in applied self-organizing systems}} (\bibinfo{year}{2013}), \bibinfo{pages}{45--55}.
\newblock


\bibitem[Da et~al\mbox{.}(2024a)]%
        {da2024cityflower}
\bibfield{author}{\bibinfo{person}{Longchao Da}, \bibinfo{person}{Chen Chu}, \bibinfo{person}{Weinan Zhang}, {and} \bibinfo{person}{Hua Wei}.} \bibinfo{year}{2024}\natexlab{a}.
\newblock \showarticletitle{CityFlowER: An Efficient and Realistic Traffic Simulator with Embedded Machine Learning Models}.
\newblock \bibinfo{journal}{\emph{arXiv preprint arXiv:2402.06127}} (\bibinfo{year}{2024}).
\newblock


\bibitem[Da et~al\mbox{.}(2024b)]%
        {da2024prompt}
\bibfield{author}{\bibinfo{person}{Longchao Da}, \bibinfo{person}{Minchiuan Gao}, \bibinfo{person}{Hao Mei}, {and} \bibinfo{person}{Hua Wei}.} \bibinfo{year}{2024}\natexlab{b}.
\newblock \showarticletitle{Prompt to transfer: Sim-to-real Transfer for Traffic Signal Control with Prompt Learning}. In \bibinfo{booktitle}{\emph{In Proceedings of the Thirty-Eighth AAAI Conference on Artificial Intelligence (AAAI'24)}}.
\newblock


\bibitem[Da et~al\mbox{.}(2024c)]%
        {da2024open}
\bibfield{author}{\bibinfo{person}{Longchao Da}, \bibinfo{person}{Kuanru Liou}, \bibinfo{person}{Tiejin Chen}, \bibinfo{person}{Xuesong Zhou}, \bibinfo{person}{Xiangyong Luo}, \bibinfo{person}{Yezhou Yang}, {and} \bibinfo{person}{Hua Wei}.} \bibinfo{year}{2024}\natexlab{c}.
\newblock \showarticletitle{Open-ti: Open traffic intelligence with augmented language model}.
\newblock \bibinfo{journal}{\emph{International Journal of Machine Learning and Cybernetics}} (\bibinfo{year}{2024}), \bibinfo{pages}{1--26}.
\newblock


\bibitem[Da et~al\mbox{.}(2023a)]%
        {da2023sim2real}
\bibfield{author}{\bibinfo{person}{Longchao Da}, \bibinfo{person}{Hao Mei}, \bibinfo{person}{Romir Sharma}, {and} \bibinfo{person}{Hua Wei}.} \bibinfo{year}{2023}\natexlab{a}.
\newblock \showarticletitle{Sim2Real Transfer for Traffic Signal Control}. In \bibinfo{booktitle}{\emph{2023 IEEE 19th International Conference on Automation Science and Engineering (CASE)}}. IEEE, \bibinfo{pages}{1--2}.
\newblock


\bibitem[Da et~al\mbox{.}(2023b)]%
        {da2023uncertainty}
\bibfield{author}{\bibinfo{person}{Longchao Da}, \bibinfo{person}{Hao Mei}, \bibinfo{person}{Romir Sharma}, {and} \bibinfo{person}{Hua Wei}.} \bibinfo{year}{2023}\natexlab{b}.
\newblock \showarticletitle{Uncertainty-aware Grounded Action Transformation towards Sim-to-Real Transfer for Traffic Signal Control}. In \bibinfo{booktitle}{\emph{In Proceedings of 62nd IEEE Conference on Decision and Control (CDC 2023)}}.
\newblock


\bibitem[Ding et~al\mbox{.}(2022)]%
        {sc4}
\bibfield{author}{\bibinfo{person}{Yi Ding}, \bibinfo{person}{Dongzhe Jiang}, \bibinfo{person}{Yu Yang}, \bibinfo{person}{Yunhuai Liu}, \bibinfo{person}{Tian He}, {and} \bibinfo{person}{Desheng Zhang}.} \bibinfo{year}{2022}\natexlab{}.
\newblock \showarticletitle{P2-Loc: {A} Person-2-Person Indoor Localization System in On-Demand Delivery}.
\newblock \bibinfo{journal}{\emph{Proc. {ACM} Interact. Mob. Wearable Ubiquitous Technol.}} \bibinfo{volume}{6}, \bibinfo{number}{1} (\bibinfo{year}{2022}), \bibinfo{pages}{9:1--9:24}.
\newblock
\urldef\tempurl%
\url{https://doi.org/10.1145/3517238}
\showDOI{\tempurl}


\bibitem[Ding et~al\mbox{.}(2020)]%
        {ding2020introduction}
\bibfield{author}{\bibinfo{person}{Zihan Ding}, \bibinfo{person}{Yanhua Huang}, \bibinfo{person}{Hang Yuan}, {and} \bibinfo{person}{Hao Dong}.} \bibinfo{year}{2020}\natexlab{}.
\newblock \showarticletitle{Introduction to reinforcement learning}.
\newblock \bibinfo{journal}{\emph{Deep reinforcement learning: fundamentals, research and applications}} (\bibinfo{year}{2020}), \bibinfo{pages}{47--123}.
\newblock


\bibitem[Du et~al\mbox{.}(2023)]%
        {du2023safelight}
\bibfield{author}{\bibinfo{person}{Wenlu Du}, \bibinfo{person}{Junyi Ye}, \bibinfo{person}{Jingyi Gu}, \bibinfo{person}{Jing Li}, \bibinfo{person}{Hua Wei}, {and} \bibinfo{person}{Guiling Wang}.} \bibinfo{year}{2023}\natexlab{}.
\newblock \showarticletitle{Safelight: A reinforcement learning method toward collision-free traffic signal control}. In \bibinfo{booktitle}{\emph{Proceedings of the AAAI Conference on Artificial Intelligence}}, Vol.~\bibinfo{volume}{37}. \bibinfo{pages}{14801--14810}.
\newblock


\bibitem[Ezell et~al\mbox{.}(2013)]%
        {ezell2013cyber}
\bibfield{author}{\bibinfo{person}{Barry~C Ezell}, \bibinfo{person}{R Michael~Robinson}, \bibinfo{person}{Peter Foytik}, \bibinfo{person}{Craig Jordan}, {and} \bibinfo{person}{David Flanagan}.} \bibinfo{year}{2013}\natexlab{}.
\newblock \showarticletitle{Cyber risk to transportation, industrial control systems, and traffic signal controllers}.
\newblock \bibinfo{journal}{\emph{Environment Systems and Decisions}}  \bibinfo{volume}{33} (\bibinfo{year}{2013}), \bibinfo{pages}{508--516}.
\newblock


\bibitem[Ghena et~al\mbox{.}(2014)]%
        {ghena2014green}
\bibfield{author}{\bibinfo{person}{Branden Ghena}, \bibinfo{person}{William Beyer}, \bibinfo{person}{Allen Hillaker}, \bibinfo{person}{Jonathan Pevarnek}, {and} \bibinfo{person}{J~Alex Halderman}.} \bibinfo{year}{2014}\natexlab{}.
\newblock \showarticletitle{Green Lights Forever: Analyzing the Security of Traffic Infrastructure.}
\newblock \bibinfo{journal}{\emph{WOOT}}  \bibinfo{volume}{14} (\bibinfo{year}{2014}), \bibinfo{pages}{7--7}.
\newblock


\bibitem[Han et~al\mbox{.}(2023)]%
        {han2023mitigating}
\bibfield{author}{\bibinfo{person}{Xiao Han}, \bibinfo{person}{Xiangyu Zhao}, \bibinfo{person}{Liang Zhang}, {and} \bibinfo{person}{Wanyu Wang}.} \bibinfo{year}{2023}\natexlab{}.
\newblock \showarticletitle{Mitigating Action Hysteresis in Traffic Signal Control with Traffic Predictive Reinforcement Learning}. In \bibinfo{booktitle}{\emph{Proceedings of the 29th ACM SIGKDD Conference on Knowledge Discovery and Data Mining}}. \bibinfo{pages}{673--684}.
\newblock


\bibitem[Hinton et~al\mbox{.}(2012)]%
        {hinton2012neural}
\bibfield{author}{\bibinfo{person}{Geoffrey Hinton}, \bibinfo{person}{Nitish Srivastava}, {and} \bibinfo{person}{Kevin Swersky}.} \bibinfo{year}{2012}\natexlab{}.
\newblock \showarticletitle{Neural networks for machine learning lecture 6a overview of mini-batch gradient descent}.
\newblock \bibinfo{journal}{\emph{Cited on}} \bibinfo{volume}{14}, \bibinfo{number}{8} (\bibinfo{year}{2012}), \bibinfo{pages}{2}.
\newblock


\bibitem[Hong et~al\mbox{.}(2022a)]%
        {sc1}
\bibfield{author}{\bibinfo{person}{Zhiqing Hong}, \bibinfo{person}{Guang Wang}, \bibinfo{person}{Wenjun Lyu}, \bibinfo{person}{Baoshen Guo}, \bibinfo{person}{Yi Ding}, \bibinfo{person}{Haotian Wang}, \bibinfo{person}{Shuai Wang}, \bibinfo{person}{Yunhuai Liu}, {and} \bibinfo{person}{Desheng Zhang}.} \bibinfo{year}{2022}\natexlab{a}.
\newblock \showarticletitle{CoMiner: nationwide behavior-driven unsupervised spatial coordinate mining from uncertain delivery events}. In \bibinfo{booktitle}{\emph{Proceedings of the 30th International Conference on Advances in Geographic Information Systems, {SIGSPATIAL} 2022, Seattle, Washington, November 1-4, 2022}}, \bibfield{editor}{\bibinfo{person}{Matthias Renz} {and} \bibinfo{person}{Mohamed Sarwat}} (Eds.). \bibinfo{publisher}{{ACM}}, \bibinfo{pages}{10:1--10:10}.
\newblock
\urldef\tempurl%
\url{https://doi.org/10.1145/3557915.3560944}
\showDOI{\tempurl}


\bibitem[Hong et~al\mbox{.}(2024)]%
        {sc8}
\bibfield{author}{\bibinfo{person}{Zhiqing Hong}, \bibinfo{person}{Haotian Wang}, \bibinfo{person}{Yi Ding}, \bibinfo{person}{Guang Wang}, \bibinfo{person}{Tian He}, {and} \bibinfo{person}{Desheng Zhang}.} \bibinfo{year}{2024}\natexlab{}.
\newblock \showarticletitle{SmallMap: Low-cost Community Road Map Sensing with Uncertain Delivery Behavior}.
\newblock \bibinfo{journal}{\emph{Proc. {ACM} Interact. Mob. Wearable Ubiquitous Technol.}} \bibinfo{volume}{8}, \bibinfo{number}{2} (\bibinfo{year}{2024}), \bibinfo{pages}{50:1--50:26}.
\newblock
\urldef\tempurl%
\url{https://doi.org/10.1145/3659596}
\showDOI{\tempurl}


\bibitem[Hong et~al\mbox{.}(2022b)]%
        {sc2}
\bibfield{author}{\bibinfo{person}{Zhiqing Hong}, \bibinfo{person}{Heng Yang}, \bibinfo{person}{Haotian Wang}, \bibinfo{person}{Wenjun Lyu}, \bibinfo{person}{Yu Yang}, \bibinfo{person}{Guang Wang}, \bibinfo{person}{Yunhuai Liu}, \bibinfo{person}{Yang Wang}, {and} \bibinfo{person}{Desheng Zhang}.} \bibinfo{year}{2022}\natexlab{b}.
\newblock \showarticletitle{FastAddr: real-time abnormal address detection via contrastive augmentation for location-based services}. In \bibinfo{booktitle}{\emph{Proceedings of the 30th International Conference on Advances in Geographic Information Systems, {SIGSPATIAL} 2022, Seattle, Washington, November 1-4, 2022}}, \bibfield{editor}{\bibinfo{person}{Matthias Renz} {and} \bibinfo{person}{Mohamed Sarwat}} (Eds.). \bibinfo{publisher}{{ACM}}, \bibinfo{pages}{64:1--64:10}.
\newblock
\urldef\tempurl%
\url{https://doi.org/10.1145/3557915.3560999}
\showDOI{\tempurl}


\bibitem[Hunter et~al\mbox{.}(2011)]%
        {hunter2011mode}
\bibfield{author}{\bibinfo{person}{Michael Hunter}, \bibinfo{person}{Peter Jenior}, \bibinfo{person}{Justin Bansen}, {and} \bibinfo{person}{Michael Rodgers}.} \bibinfo{year}{2011}\natexlab{}.
\newblock \showarticletitle{Mode of flashing for malfunctioning traffic signals}.
\newblock \bibinfo{journal}{\emph{Journal of transportation engineering}} \bibinfo{volume}{137}, \bibinfo{number}{7} (\bibinfo{year}{2011}), \bibinfo{pages}{438--444}.
\newblock


\bibitem[Jiang et~al\mbox{.}(2024a)]%
        {jiang2024x}
\bibfield{author}{\bibinfo{person}{Haoyuan Jiang}, \bibinfo{person}{Ziyue Li}, \bibinfo{person}{Hua Wei}, \bibinfo{person}{Xuantang Xiong}, \bibinfo{person}{Jingqing Ruan}, \bibinfo{person}{Jiaming Lu}, \bibinfo{person}{Hangyu Mao}, {and} \bibinfo{person}{Rui Zhao}.} \bibinfo{year}{2024}\natexlab{a}.
\newblock \showarticletitle{X-Light: Cross-City Traffic Signal Control Using Transformer on Transformer as Meta Multi-Agent Reinforcement Learner}.
\newblock \bibinfo{journal}{\emph{arXiv preprint arXiv:2404.12090}} (\bibinfo{year}{2024}).
\newblock


\bibitem[Jiang et~al\mbox{.}(2024b)]%
        {jiang2024guidelight}
\bibfield{author}{\bibinfo{person}{Haoyuan Jiang}, \bibinfo{person}{Xuantang Xiong}, \bibinfo{person}{Ziyue Li}, \bibinfo{person}{Hangyu Mao}, \bibinfo{person}{Guanghu Sui}, \bibinfo{person}{Jingqing Ruan}, \bibinfo{person}{Yuheng Cheng}, \bibinfo{person}{Hua Wei}, \bibinfo{person}{Wolfgang Ketter}, {and} \bibinfo{person}{Rui Zhao}.} \bibinfo{year}{2024}\natexlab{b}.
\newblock \showarticletitle{GuideLight:" Industrial Solution" Guidance for More Practical Traffic Signal Control Agents}.
\newblock \bibinfo{journal}{\emph{arXiv preprint arXiv:2407.10811}} (\bibinfo{year}{2024}).
\newblock


\bibitem[Koonce and Rodegerdts(2008)]%
        {koonce2008traffic}
\bibfield{author}{\bibinfo{person}{Peter Koonce} {and} \bibinfo{person}{Lee Rodegerdts}.} \bibinfo{year}{2008}\natexlab{}.
\newblock \bibinfo{booktitle}{\emph{Traffic signal timing manual.}}
\newblock \bibinfo{type}{{T}echnical {R}eport}. \bibinfo{institution}{United States. Federal Highway Administration}.
\newblock


\bibitem[Kuppusamy et~al\mbox{.}(2018)]%
        {kuppusamy2018design}
\bibfield{author}{\bibinfo{person}{P Kuppusamy}, \bibinfo{person}{P Kamarajapandian}, \bibinfo{person}{MS Sabari}, {and} \bibinfo{person}{J Nithya}.} \bibinfo{year}{2018}\natexlab{}.
\newblock \showarticletitle{Design of smart traffic signal system using internet of things and genetic algorithm}. In \bibinfo{booktitle}{\emph{Advances in Big Data and Cloud Computing}}. Springer, \bibinfo{pages}{395--403}.
\newblock


\bibitem[Lee et~al\mbox{.}(2019)]%
        {lee2019cyber}
\bibfield{author}{\bibinfo{person}{JooChan Lee}, \bibinfo{person}{JangHoon Kim}, {and} \bibinfo{person}{JungTaek Seo}.} \bibinfo{year}{2019}\natexlab{}.
\newblock \showarticletitle{Cyber attack scenarios on smart city and their ripple effects}. In \bibinfo{booktitle}{\emph{2019 international conference on platform technology and service (PlatCon)}}. IEEE, \bibinfo{pages}{1--5}.
\newblock


\bibitem[Lhomme et~al\mbox{.}(2013)]%
        {lhomme2013analyzing}
\bibfield{author}{\bibinfo{person}{Serge Lhomme}, \bibinfo{person}{Damien Serre}, \bibinfo{person}{Youssef Diab}, {and} \bibinfo{person}{Richard Laganier}.} \bibinfo{year}{2013}\natexlab{}.
\newblock \showarticletitle{Analyzing resilience of urban networks: a preliminary step towards more flood resilient cities}.
\newblock \bibinfo{journal}{\emph{Natural hazards and earth system sciences}} \bibinfo{volume}{13}, \bibinfo{number}{2} (\bibinfo{year}{2013}), \bibinfo{pages}{221--230}.
\newblock


\bibitem[Li et~al\mbox{.}(2018)]%
        {li2017diffusion}
\bibfield{author}{\bibinfo{person}{Yaguang Li}, \bibinfo{person}{Rose Yu}, \bibinfo{person}{Cyrus Shahabi}, {and} \bibinfo{person}{Yan Liu}.} \bibinfo{year}{2018}\natexlab{}.
\newblock \showarticletitle{Diffusion Convolutional Recurrent Neural Network: Data-Driven Traffic Forecasting}. In \bibinfo{booktitle}{\emph{6th International Conference on Learning Representations, {ICLR} 2018, Vancouver, BC, Canada, April 30 - May 3, 2018, Conference Track Proceedings}}. \bibinfo{publisher}{OpenReview.net}.
\newblock
\urldef\tempurl%
\url{https://openreview.net/forum?id=SJiHXGWAZ}
\showURL{%
\tempurl}


\bibitem[Lopez et~al\mbox{.}(2018)]%
        {SUMO2018}
\bibfield{author}{\bibinfo{person}{Pablo~Alvarez Lopez}, \bibinfo{person}{Michael Behrisch}, \bibinfo{person}{Laura Bieker-Walz}, \bibinfo{person}{Jakob Erdmann}, \bibinfo{person}{Yun-Pang Fl{\"o}tter{\"o}d}, \bibinfo{person}{Robert Hilbrich}, \bibinfo{person}{Leonhard L{\"u}cken}, \bibinfo{person}{Johannes Rummel}, \bibinfo{person}{Peter Wagner}, {and} \bibinfo{person}{Evamarie Wie{\ss}ner}.} \bibinfo{year}{2018}\natexlab{}.
\newblock \showarticletitle{Microscopic Traffic Simulation using SUMO}, In \bibinfo{booktitle}{The 21st IEEE International Conference on Intelligent Transportation Systems}.
\newblock \bibinfo{journal}{\emph{IEEE Intelligent Transportation Systems Conference (ITSC)}}.
\newblock
\urldef\tempurl%
\url{https://elib.dlr.de/124092/}
\showURL{%
\tempurl}


\bibitem[Lyu et~al\mbox{.}(2023)]%
        {rl2}
\bibfield{author}{\bibinfo{person}{Wenjun Lyu}, \bibinfo{person}{Haotian Wang}, \bibinfo{person}{Zhiqing Hong}, \bibinfo{person}{Guang Wang}, \bibinfo{person}{Yu Yang}, \bibinfo{person}{Yunhuai Liu}, {and} \bibinfo{person}{Desheng Zhang}.} \bibinfo{year}{2023}\natexlab{}.
\newblock \showarticletitle{{REDE:} Exploring Relay Transportation for Efficient Last-mile Delivery}. In \bibinfo{booktitle}{\emph{39th {IEEE} International Conference on Data Engineering, {ICDE} 2023, Anaheim, CA, USA, April 3-7, 2023}}. \bibinfo{publisher}{{IEEE}}, \bibinfo{pages}{3003--3016}.
\newblock
\urldef\tempurl%
\url{https://doi.org/10.1109/ICDE55515.2023.00230}
\showDOI{\tempurl}


\bibitem[Lyu et~al\mbox{.}(2022)]%
        {rl3}
\bibfield{author}{\bibinfo{person}{Wenjun Lyu}, \bibinfo{person}{Kexin Zhang}, \bibinfo{person}{Baoshen Guo}, \bibinfo{person}{Zhiqing Hong}, \bibinfo{person}{Guang Yang}, \bibinfo{person}{Guang Wang}, \bibinfo{person}{Yu Yang}, \bibinfo{person}{Yunhuai Liu}, {and} \bibinfo{person}{Desheng Zhang}.} \bibinfo{year}{2022}\natexlab{}.
\newblock \showarticletitle{Towards Fair Workload Assessment via Homogeneous Order Grouping in Last-mile Delivery}. In \bibinfo{booktitle}{\emph{Proceedings of the 31st {ACM} International Conference on Information {\&} Knowledge Management, Atlanta, GA, USA, October 17-21, 2022}}, \bibfield{editor}{\bibinfo{person}{Mohammad~Al Hasan} {and} \bibinfo{person}{Li~Xiong}} (Eds.). \bibinfo{publisher}{{ACM}}, \bibinfo{pages}{3361--3370}.
\newblock
\urldef\tempurl%
\url{https://doi.org/10.1145/3511808.3557132}
\showDOI{\tempurl}


\bibitem[Mathibela(2017)]%
        {mathibela2017another}
\bibfield{author}{\bibinfo{person}{Bonolo Mathibela}.} \bibinfo{year}{2017}\natexlab{}.
\newblock \showarticletitle{Another broken traffic light? Reducing traffic congestion in resource constrained environments}. In \bibinfo{booktitle}{\emph{2017 IEEE Intelligent Vehicles Symposium (IV)}}. IEEE, \bibinfo{pages}{1305--1310}.
\newblock


\bibitem[Mei et~al\mbox{.}(2022)]%
        {mei2022libsignal}
\bibfield{author}{\bibinfo{person}{Hao Mei}, \bibinfo{person}{Xiaoliang Lei}, \bibinfo{person}{Longchao Da}, \bibinfo{person}{Bin Shi}, {and} \bibinfo{person}{Hua Wei}.} \bibinfo{year}{2022}\natexlab{}.
\newblock \showarticletitle{LibSignal: An Open Library for Traffic Signal Control}.
\newblock \bibinfo{journal}{\emph{arXiv preprint arXiv:2211.10649}} (\bibinfo{year}{2022}).
\newblock


\bibitem[Mei et~al\mbox{.}(2023)]%
        {mei2023reinforcement}
\bibfield{author}{\bibinfo{person}{Hao Mei}, \bibinfo{person}{Junxian Li}, \bibinfo{person}{Bin Shi}, {and} \bibinfo{person}{Hua Wei}.} \bibinfo{year}{2023}\natexlab{}.
\newblock \showarticletitle{Reinforcement Learning Approaches for Traffic Signal Control under Missing Data}.
\newblock \bibinfo{journal}{\emph{arXiv preprint arXiv:2304.10722}} (\bibinfo{year}{2023}).
\newblock


\bibitem[{National Center for Statistics and Analysis}(2022)]%
        {NCSA2022}
\bibfield{author}{\bibinfo{person}{{National Center for Statistics and Analysis}}.} \bibinfo{year}{2022}\natexlab{}.
\newblock \bibinfo{booktitle}{\emph{Traffic Safety Facts 2020: A Compilation of Motor Vehicle Crash Data}}.
\newblock \bibinfo{type}{Report} DOT HS 813 375. \bibinfo{institution}{National Highway Traffic Safety Administration}.
\newblock


\bibitem[Ng et~al\mbox{.}(1999)]%
        {ng1999policy}
\bibfield{author}{\bibinfo{person}{Andrew~Y Ng}, \bibinfo{person}{Daishi Harada}, {and} \bibinfo{person}{Stuart Russell}.} \bibinfo{year}{1999}\natexlab{}.
\newblock \showarticletitle{Policy invariance under reward transformations: Theory and application to reward shaping}. In \bibinfo{booktitle}{\emph{Icml}}, Vol.~\bibinfo{volume}{99}. Citeseer, \bibinfo{pages}{278--287}.
\newblock


\bibitem[Paul and Pitale(2017)]%
        {paul2017comparison}
\bibfield{author}{\bibinfo{person}{Basil Paul} {and} \bibinfo{person}{H Pitale}.} \bibinfo{year}{2017}\natexlab{}.
\newblock \showarticletitle{Comparison of capacity at signalised and unsignalised intersection}.
\newblock \bibinfo{journal}{\emph{IJSTE-Int J Sci Technol Eng}}  \bibinfo{volume}{3} (\bibinfo{year}{2017}), \bibinfo{pages}{108--111}.
\newblock


\bibitem[Pregnolato et~al\mbox{.}(2016)]%
        {pregnolato2016assessing}
\bibfield{author}{\bibinfo{person}{Maria Pregnolato}, \bibinfo{person}{Alistair Ford}, \bibinfo{person}{Craig Robson}, \bibinfo{person}{Vassilis Glenis}, \bibinfo{person}{Stuart Barr}, {and} \bibinfo{person}{Richard Dawson}.} \bibinfo{year}{2016}\natexlab{}.
\newblock \showarticletitle{Assessing urban strategies for reducing the impacts of extreme weather on infrastructure networks}.
\newblock \bibinfo{journal}{\emph{Royal Society open science}} \bibinfo{volume}{3}, \bibinfo{number}{5} (\bibinfo{year}{2016}), \bibinfo{pages}{160023}.
\newblock


\bibitem[Purba et~al\mbox{.}(2018)]%
        {purba2018developing}
\bibfield{author}{\bibinfo{person}{Aleksander Purba}, \bibinfo{person}{Rahayu Sulistyorini}, {and} \bibinfo{person}{Ageng Sadnowo}.} \bibinfo{year}{2018}\natexlab{}.
\newblock \showarticletitle{Developing Monitoring System of Traffic Signal Using Microcontroller Device by SMS of GSM Network}.
\newblock \bibinfo{journal}{\emph{Prosiding Semnas SINTA FT UNILA Vol. 1 Tahun 2018}} \bibinfo{volume}{1}, \bibinfo{number}{1} (\bibinfo{year}{2018}), \bibinfo{pages}{273--277}.
\newblock


\bibitem[Rodrigues and Azevedo(2019)]%
        {rodrigues2019towards}
\bibfield{author}{\bibinfo{person}{Filipe Rodrigues} {and} \bibinfo{person}{Carlos~Lima Azevedo}.} \bibinfo{year}{2019}\natexlab{}.
\newblock \showarticletitle{Towards robust deep reinforcement learning for traffic signal control: Demand surges, incidents and sensor failures}. In \bibinfo{booktitle}{\emph{2019 IEEE intelligent transportation systems conference (ITSC)}}. IEEE, \bibinfo{pages}{3559--3566}.
\newblock


\bibitem[Rosenbloom(1978)]%
        {rosenbloom1978peak}
\bibfield{author}{\bibinfo{person}{Sandra Rosenbloom}.} \bibinfo{year}{1978}\natexlab{}.
\newblock \showarticletitle{Peak-period traffic congestion: A state-of-the-art analysis and evaluation of effective solutions}.
\newblock \bibinfo{journal}{\emph{Transportation}} \bibinfo{volume}{7}, \bibinfo{number}{2} (\bibinfo{year}{1978}), \bibinfo{pages}{167--191}.
\newblock


\bibitem[Ruan et~al\mbox{.}(2024)]%
        {ruan2024coslight}
\bibfield{author}{\bibinfo{person}{Jingqing Ruan}, \bibinfo{person}{Ziyue Li}, \bibinfo{person}{Hua Wei}, \bibinfo{person}{Haoyuan Jiang}, \bibinfo{person}{Jiaming Lu}, \bibinfo{person}{Xuantang Xiong}, \bibinfo{person}{Hangyu Mao}, {and} \bibinfo{person}{Rui Zhao}.} \bibinfo{year}{2024}\natexlab{}.
\newblock \showarticletitle{CoSLight: Co-optimizing Collaborator Selection and Decision-making to Enhance Traffic Signal Control}.
\newblock \bibinfo{journal}{\emph{arXiv preprint arXiv:2405.17152}} (\bibinfo{year}{2024}).
\newblock


\bibitem[Shuman et~al\mbox{.}(2013)]%
        {shuman2013emerging}
\bibfield{author}{\bibinfo{person}{David~I Shuman}, \bibinfo{person}{Sunil~K Narang}, \bibinfo{person}{Pascal Frossard}, \bibinfo{person}{Antonio Ortega}, {and} \bibinfo{person}{Pierre Vandergheynst}.} \bibinfo{year}{2013}\natexlab{}.
\newblock \showarticletitle{The emerging field of signal processing on graphs: Extending high-dimensional data analysis to networks and other irregular domains}.
\newblock \bibinfo{journal}{\emph{IEEE signal processing magazine}} \bibinfo{volume}{30}, \bibinfo{number}{3} (\bibinfo{year}{2013}), \bibinfo{pages}{83--98}.
\newblock


\bibitem[Soh et~al\mbox{.}(2013)]%
        {soh2013smart}
\bibfield{author}{\bibinfo{person}{Azura~Che Soh}, \bibinfo{person}{Asnor~Juraiza Ishak}, {and} \bibinfo{person}{Mohd~Hanif Zaini}.} \bibinfo{year}{2013}\natexlab{}.
\newblock \showarticletitle{Smart monitoring fault detection system for malfunction traffic light operation}. In \bibinfo{booktitle}{\emph{2013 IEEE 8th Conference on Industrial Electronics and Applications (ICIEA)}}. IEEE, \bibinfo{pages}{549--554}.
\newblock


\bibitem[Stevanovic et~al\mbox{.}(2009)]%
        {stevanovic2009scoot}
\bibfield{author}{\bibinfo{person}{Aleksandar Stevanovic}, \bibinfo{person}{Cameron Kergaye}, {and} \bibinfo{person}{Peter~T Martin}.} \bibinfo{year}{2009}\natexlab{}.
\newblock \showarticletitle{Scoot and scats: A closer look into their operations}. In \bibinfo{booktitle}{\emph{88th Annual Meeting of the Transportation Research Board. Washington DC}}.
\newblock


\bibitem[Suarez et~al\mbox{.}(2005)]%
        {suarez2005impacts}
\bibfield{author}{\bibinfo{person}{Pablo Suarez}, \bibinfo{person}{William Anderson}, \bibinfo{person}{Vijay Mahal}, {and} \bibinfo{person}{TR Lakshmanan}.} \bibinfo{year}{2005}\natexlab{}.
\newblock \showarticletitle{Impacts of flooding and climate change on urban transportation: A systemwide performance assessment of the Boston Metro Area}.
\newblock \bibinfo{journal}{\emph{Transportation Research Part D: transport and environment}} \bibinfo{volume}{10}, \bibinfo{number}{3} (\bibinfo{year}{2005}), \bibinfo{pages}{231--244}.
\newblock


\bibitem[Taale et~al\mbox{.}(2004)]%
        {taale2004evaluation}
\bibfield{author}{\bibinfo{person}{Henk Taale}, \bibinfo{person}{PHG van Bekkum}, {and} \bibinfo{person}{MLD van Rij}.} \bibinfo{year}{2004}\natexlab{}.
\newblock \showarticletitle{Evaluation of traffic management by the traffic police}.
\newblock  (\bibinfo{year}{2004}).
\newblock


\bibitem[Tan et~al\mbox{.}(2023)]%
        {rl1}
\bibfield{author}{\bibinfo{person}{Heng Tan}, \bibinfo{person}{Yukun Yuan}, \bibinfo{person}{Shuxin Zhong}, {and} \bibinfo{person}{Yu Yang}.} \bibinfo{year}{2023}\natexlab{}.
\newblock \showarticletitle{Joint Rebalancing and Charging for Shared Electric Micromobility Vehicles with Energy-informed Demand}. In \bibinfo{booktitle}{\emph{Proceedings of the 32nd {ACM} International Conference on Information and Knowledge Management, {CIKM} 2023, Birmingham, United Kingdom, October 21-25, 2023}}, \bibfield{editor}{\bibinfo{person}{Ingo Frommholz}, \bibinfo{person}{Frank Hopfgartner}, \bibinfo{person}{Mark Lee}, \bibinfo{person}{Michael Oakes}, \bibinfo{person}{Mounia Lalmas}, \bibinfo{person}{Min Zhang}, {and} \bibinfo{person}{Rodrygo L.~T. Santos}} (Eds.). \bibinfo{publisher}{{ACM}}, \bibinfo{pages}{2392--2401}.
\newblock
\urldef\tempurl%
\url{https://doi.org/10.1145/3583780.3614942}
\showDOI{\tempurl}


\bibitem[Teng et~al\mbox{.}(2016)]%
        {teng2016scalable}
\bibfield{author}{\bibinfo{person}{Shang-Hua Teng} {et~al\mbox{.}}} \bibinfo{year}{2016}\natexlab{}.
\newblock \showarticletitle{Scalable algorithms for data and network analysis}.
\newblock \bibinfo{journal}{\emph{Foundations and Trends{\textregistered} in Theoretical Computer Science}} \bibinfo{volume}{12}, \bibinfo{number}{1--2} (\bibinfo{year}{2016}), \bibinfo{pages}{1--274}.
\newblock


\bibitem[Varaiya(2013)]%
        {varaiya2013max}
\bibfield{author}{\bibinfo{person}{Pravin Varaiya}.} \bibinfo{year}{2013}\natexlab{}.
\newblock \showarticletitle{Max pressure control of a network of signalized intersections}.
\newblock \bibinfo{journal}{\emph{Transportation Research Part C: Emerging Technologies}}  \bibinfo{volume}{36} (\bibinfo{year}{2013}), \bibinfo{pages}{177--195}.
\newblock


\bibitem[Vlachogiannis et~al\mbox{.}(2023)]%
        {vlachogiannis2023humanlight}
\bibfield{author}{\bibinfo{person}{Dimitris~M Vlachogiannis}, \bibinfo{person}{Hua Wei}, \bibinfo{person}{Scott Moura}, {and} \bibinfo{person}{Jane Macfarlane}.} \bibinfo{year}{2023}\natexlab{}.
\newblock \showarticletitle{HumanLight: Incentivizing ridesharing via human-centric deep reinforcement learning in traffic signal control}.
\newblock \bibinfo{journal}{\emph{Transportation Research Part C: Emerging Technologies}} (\bibinfo{year}{2023}).
\newblock


\bibitem[Wei et~al\mbox{.}(2019a)]%
        {wei2019presslight}
\bibfield{author}{\bibinfo{person}{Hua Wei}, \bibinfo{person}{Chacha Chen}, \bibinfo{person}{Guanjie Zheng}, \bibinfo{person}{Kan Wu}, \bibinfo{person}{Vikash Gayah}, \bibinfo{person}{Kai Xu}, {and} \bibinfo{person}{Zhenhui Li}.} \bibinfo{year}{2019}\natexlab{a}.
\newblock \showarticletitle{Presslight: Learning max pressure control to coordinate traffic signals in arterial network}. In \bibinfo{booktitle}{\emph{Proceedings of the 25th ACM SIGKDD International Conference on Knowledge Discovery \& Data Mining}}. \bibinfo{pages}{1290--1298}.
\newblock


\bibitem[Wei et~al\mbox{.}(2019b)]%
        {wei2019colight}
\bibfield{author}{\bibinfo{person}{Hua Wei}, \bibinfo{person}{Nan Xu}, \bibinfo{person}{Huichu Zhang}, \bibinfo{person}{Guanjie Zheng}, \bibinfo{person}{Xinshi Zang}, \bibinfo{person}{Chacha Chen}, \bibinfo{person}{Weinan Zhang}, \bibinfo{person}{Yanmin Zhu}, \bibinfo{person}{Kai Xu}, {and} \bibinfo{person}{Zhenhui Li}.} \bibinfo{year}{2019}\natexlab{b}.
\newblock \showarticletitle{Colight: Learning network-level cooperation for traffic signal control}. In \bibinfo{booktitle}{\emph{Proceedings of the 28th ACM International Conference on Information and Knowledge Management}}. \bibinfo{pages}{1913--1922}.
\newblock


\bibitem[Wei et~al\mbox{.}(2019c)]%
        {wei2019survey}
\bibfield{author}{\bibinfo{person}{Hua Wei}, \bibinfo{person}{Guanjie Zheng}, \bibinfo{person}{Vikash Gayah}, {and} \bibinfo{person}{Zhenhui Li}.} \bibinfo{year}{2019}\natexlab{c}.
\newblock \showarticletitle{A survey on traffic signal control methods}.
\newblock \bibinfo{journal}{\emph{arXiv preprint arXiv:1904.08117}} (\bibinfo{year}{2019}).
\newblock


\bibitem[Wei et~al\mbox{.}(2018)]%
        {wei2018intellilight}
\bibfield{author}{\bibinfo{person}{Hua Wei}, \bibinfo{person}{Guanjie Zheng}, \bibinfo{person}{Huaxiu Yao}, {and} \bibinfo{person}{Zhenhui Li}.} \bibinfo{year}{2018}\natexlab{}.
\newblock \showarticletitle{Intellilight: A reinforcement learning approach for intelligent traffic light control}. In \bibinfo{booktitle}{\emph{Proceedings of the 24th ACM SIGKDD International Conference on Knowledge Discovery \& Data Mining}}. \bibinfo{pages}{2496--2505}.
\newblock


\bibitem[Wu et~al\mbox{.}(2021)]%
        {wu2021dynstgat}
\bibfield{author}{\bibinfo{person}{Libing Wu}, \bibinfo{person}{Min Wang}, \bibinfo{person}{Dan Wu}, {and} \bibinfo{person}{Jia Wu}.} \bibinfo{year}{2021}\natexlab{}.
\newblock \showarticletitle{DynSTGAT: Dynamic spatial-temporal graph attention network for traffic signal control}. In \bibinfo{booktitle}{\emph{Proceedings of the 30th ACM International Conference on Information \& Knowledge Management}}. \bibinfo{pages}{2150--2159}.
\newblock


\bibitem[Yan et~al\mbox{.}(2024)]%
        {sc7}
\bibfield{author}{\bibinfo{person}{Hua Yan}, \bibinfo{person}{Heng Tan}, \bibinfo{person}{Haotian Wang}, \bibinfo{person}{Desheng Zhang}, {and} \bibinfo{person}{Yu Yang}.} \bibinfo{year}{2024}\natexlab{}.
\newblock \showarticletitle{Robust Route Planning under Uncertain Pickup Requests for Last-mile Delivery}. In \bibinfo{booktitle}{\emph{Proceedings of the {ACM} on Web Conference 2024, {WWW} 2024, Singapore, May 13-17, 2024}}, \bibfield{editor}{\bibinfo{person}{Tat{-}Seng Chua}, \bibinfo{person}{Chong{-}Wah Ngo}, \bibinfo{person}{Ravi Kumar}, \bibinfo{person}{Hady~W. Lauw}, {and} \bibinfo{person}{Roy~Ka{-}Wei Lee}} (Eds.). \bibinfo{publisher}{{ACM}}, \bibinfo{pages}{3022--3030}.
\newblock
\urldef\tempurl%
\url{https://doi.org/10.1145/3589334.3645595}
\showDOI{\tempurl}


\bibitem[Yang et~al\mbox{.}(2023)]%
        {sc5}
\bibfield{author}{\bibinfo{person}{Guang Yang}, \bibinfo{person}{Yuequn Zhang}, \bibinfo{person}{Jinquan Hang}, \bibinfo{person}{Xinyue Feng}, \bibinfo{person}{Zejun Xie}, \bibinfo{person}{Desheng Zhang}, {and} \bibinfo{person}{Yu Yang}.} \bibinfo{year}{2023}\natexlab{}.
\newblock \showarticletitle{{CARPG:} Cross-City Knowledge Transfer for Traffic Accident Prediction via Attentive Region-Level Parameter Generation}. In \bibinfo{booktitle}{\emph{Proceedings of the 32nd {ACM} International Conference on Information and Knowledge Management, {CIKM} 2023, Birmingham, United Kingdom, October 21-25, 2023}}, \bibfield{editor}{\bibinfo{person}{Ingo Frommholz}, \bibinfo{person}{Frank Hopfgartner}, \bibinfo{person}{Mark Lee}, \bibinfo{person}{Michael Oakes}, \bibinfo{person}{Mounia Lalmas}, \bibinfo{person}{Min Zhang}, {and} \bibinfo{person}{Rodrygo L.~T. Santos}} (Eds.). \bibinfo{publisher}{{ACM}}, \bibinfo{pages}{2939--2948}.
\newblock
\urldef\tempurl%
\url{https://doi.org/10.1145/3583780.3614802}
\showDOI{\tempurl}


\bibitem[Yang et~al\mbox{.}(2019)]%
        {sc3}
\bibfield{author}{\bibinfo{person}{Qinchen Yang}, \bibinfo{person}{Man Liu}, \bibinfo{person}{Zhitao Zhang}, \bibinfo{person}{Shuqin Yang}, \bibinfo{person}{Jifeng Ning}, {and} \bibinfo{person}{Wenting Han}.} \bibinfo{year}{2019}\natexlab{}.
\newblock \showarticletitle{Mapping Plastic Mulched Farmland for High Resolution Images of Unmanned Aerial Vehicle Using Deep Semantic Segmentation}.
\newblock \bibinfo{journal}{\emph{Remote. Sens.}} \bibinfo{volume}{11}, \bibinfo{number}{17} (\bibinfo{year}{2019}), \bibinfo{pages}{2008}.
\newblock
\urldef\tempurl%
\url{https://doi.org/10.3390/RS11172008}
\showDOI{\tempurl}


\bibitem[Zhang et~al\mbox{.}(2019)]%
        {zhang2019cityflow}
\bibfield{author}{\bibinfo{person}{Huichu Zhang}, \bibinfo{person}{Siyuan Feng}, \bibinfo{person}{Chang Liu}, \bibinfo{person}{Yaoyao Ding}, \bibinfo{person}{Yichen Zhu}, \bibinfo{person}{Zihan Zhou}, \bibinfo{person}{Weinan Zhang}, \bibinfo{person}{Yong Yu}, \bibinfo{person}{Haiming Jin}, {and} \bibinfo{person}{Zhenhui Li}.} \bibinfo{year}{2019}\natexlab{}.
\newblock \showarticletitle{Cityflow: A multi-agent reinforcement learning environment for large scale city traffic scenario}. In \bibinfo{booktitle}{\emph{The world wide web conference}}. \bibinfo{pages}{3620--3624}.
\newblock


\bibitem[Zhang et~al\mbox{.}(2022)]%
        {zhang2022expression}
\bibfield{author}{\bibinfo{person}{Liang Zhang}, \bibinfo{person}{Qiang Wu}, \bibinfo{person}{Jun Shen}, \bibinfo{person}{Linyuan L{\"u}}, \bibinfo{person}{Bo Du}, {and} \bibinfo{person}{Jianqing Wu}.} \bibinfo{year}{2022}\natexlab{}.
\newblock \showarticletitle{Expression might be enough: Representing pressure and demand for reinforcement learning based traffic signal control}. In \bibinfo{booktitle}{\emph{International Conference on Machine Learning}}. PMLR, \bibinfo{pages}{26645--26654}.
\newblock


\bibitem[Zheng et~al\mbox{.}(2019)]%
        {zheng2019learning}
\bibfield{author}{\bibinfo{person}{Guanjie Zheng}, \bibinfo{person}{Yuanhao Xiong}, \bibinfo{person}{Xinshi Zang}, \bibinfo{person}{Jie Feng}, \bibinfo{person}{Hua Wei}, \bibinfo{person}{Huichu Zhang}, \bibinfo{person}{Yong Li}, \bibinfo{person}{Kai Xu}, {and} \bibinfo{person}{Zhenhui Li}.} \bibinfo{year}{2019}\natexlab{}.
\newblock \showarticletitle{Learning phase competition for traffic signal control}. In \bibinfo{booktitle}{\emph{Proceedings of the 28th ACM International Conference on Information and Knowledge Management}}. \bibinfo{pages}{1963--1972}.
\newblock


\bibitem[Zhong et~al\mbox{.}(2023)]%
        {sc6}
\bibfield{author}{\bibinfo{person}{Shuxin Zhong}, \bibinfo{person}{William Yubeaton}, \bibinfo{person}{Wenjun Lyu}, \bibinfo{person}{Guang Wang}, \bibinfo{person}{Desheng Zhang}, {and} \bibinfo{person}{Yu Yang}.} \bibinfo{year}{2023}\natexlab{}.
\newblock \showarticletitle{{RLIFE:} Remaining Lifespan Prediction for E-scooters}. In \bibinfo{booktitle}{\emph{Proceedings of the 32nd {ACM} International Conference on Information and Knowledge Management, {CIKM} 2023, Birmingham, United Kingdom, October 21-25, 2023}}, \bibfield{editor}{\bibinfo{person}{Ingo Frommholz}, \bibinfo{person}{Frank Hopfgartner}, \bibinfo{person}{Mark Lee}, \bibinfo{person}{Michael Oakes}, \bibinfo{person}{Mounia Lalmas}, \bibinfo{person}{Min Zhang}, {and} \bibinfo{person}{Rodrygo L.~T. Santos}} (Eds.). \bibinfo{publisher}{{ACM}}, \bibinfo{pages}{3544--3553}.
\newblock
\urldef\tempurl%
\url{https://doi.org/10.1145/3583780.3615037}
\showDOI{\tempurl}


\end{thebibliography}

\end{document}